\documentclass{article}
\usepackage{fontenc}   
\usepackage{wrapfig}
\usepackage{amssymb}
\usepackage{amsmath}
\usepackage{booktabs}
\usepackage{enumitem}
\usepackage{graphicx}
\usepackage{color, soul}
\usepackage{xcolor} 
\usepackage{booktabs}
\usepackage{colortbl}
\usepackage{tcolorbox}
\usepackage{color,xcolor}
\usepackage{microtype}
\usepackage{inconsolata}
\usepackage{fancyvrb}
\usepackage{multirow}
\usepackage{colortbl}
\usepackage{lipsum}
\usepackage{pgfplots}
\usepackage{siunitx}
\usepackage{hyperref}
\usepackage{subcaption}
\usepackage{float} 
\usepackage{hyperref}

\usepackage{setspace} 
\usepackage{diagbox} 
\usepackage{xcolor}
\usepackage{hyperref}

\usepackage[accepted]{icml2024}
\usepackage{amsmath}
\usepackage{amssymb}
\usepackage{mathtools}
\usepackage{amsthm}
\usepackage{microtype}
\usepackage{graphicx}
\usepackage{subcaption}
\usepackage{lipsum}
\usepackage{array} 
\usepackage{etex}
\usepackage{adjustbox}
\usepackage{pifont}

\usepackage[capitalize,noabbrev]{cleveref}

\theoremstyle{plain}

\theoremstyle{definition}

\theoremstyle{remark}

\usepackage[textsize=tiny]{todonotes}
\usepackage{float} 
\usepackage{makecell}   

\usepackage{minitoc} 

\icmltitlerunning{BioMamba: Leveraging Spectro-Temporal Embedding in Bidirectional Mamba for Enhanced Biosignal Classification}

\begin{document}

\newcommand{\deepred}[1]{\textcolor{red!60!black}{\mathversion{bold}#1}}
\newcommand{\deepreds}[1]{\mathversion{bold}#1}
\newcommand{\deepblue}[1]{\textcolor{blue!60!black}{\mathversion{bold}#1}}
\renewcommand{\baselinestretch}{}
\newcommand{\cyanb}[1]{\colorbox{cyan!20}{#1}}

\twocolumn[
\icmltitle{BioMamba: Leveraging Spectro-Temporal Embedding in Bidirectional Mamba \\ for Enhanced Biosignal Classification}

\icmlsetsymbol{equal}{*}

\begin{icmlauthorlist}
\icmlauthor{Jian Qian}{}
\icmlauthor{Teck Lun Goh}{}
\icmlauthor{Bingyu Xie}{}
\icmlauthor{Chengyao Zhu}{}
\icmlauthor{Biao Wan}{}
\icmlauthor{Yawen Guan}{}
\icmlauthor{Rachel Ding Chen}{}
\icmlauthor{Patrick Yin Chiang}{yyy}
\end{icmlauthorlist}
\icmlaffiliation{yyy}{Fudan University}
\icmlcorrespondingauthor{Patrick Yin Chiang}{pchiang@fudan.edu.cn}
\icmlkeywords{Machine Learning, ICML}
\vskip 0.3in
]
\printAffiliationsAndNotice{}  

\begin{abstract}
Biological signals, such as electroencephalograms (EEGs) and electrocardiograms (ECGs), play a pivotal role in numerous clinical practices, such as diagnosing
brain and cardiac arrhythmic diseases.  Existing methods for biosignal classification rely on Attention-based frameworks with dense Feed Forward layers, which lead to inefficient learning, high computational overhead, and suboptimal performance. In this work, we introduce~\textbf{BioMamba}, a ~\textbf{Spectro-Temporal Embedding} strategy applied to the~\textbf{Bidirectional Mamba} framework with~\textbf{Sparse Feed Forward} layers to enable effective learning of biosignal sequences. By integrating these three key components, BioMamba effectively addresses the limitations of existing methods.
Extensive experiments demonstrate that~{BioMamba} significantly outperforms state-of-the-art methods with marked improvement in classification performance. The advantages of the proposed~{BioMamba} include 
(1)~\textbf{Reliability:} BioMamba consistently delivers robust results, confirmed across six evaluation metrics.
(2)~\textbf{Efficiency:} We assess both model and training efficiency, the BioMamba demonstrates computational effectiveness by reducing model size and resource consumption compared to existing approaches.
(3)~\textbf{Generality:} With the capacity to effectively classify a diverse set of tasks, BioMamba demonstrates adaptability and effectiveness across various domains and applications. 
\end{abstract}

\section{Introduction}
\label{sec: intro}

\begin{figure}[t]
\centering
\includegraphics[width=1\columnwidth]{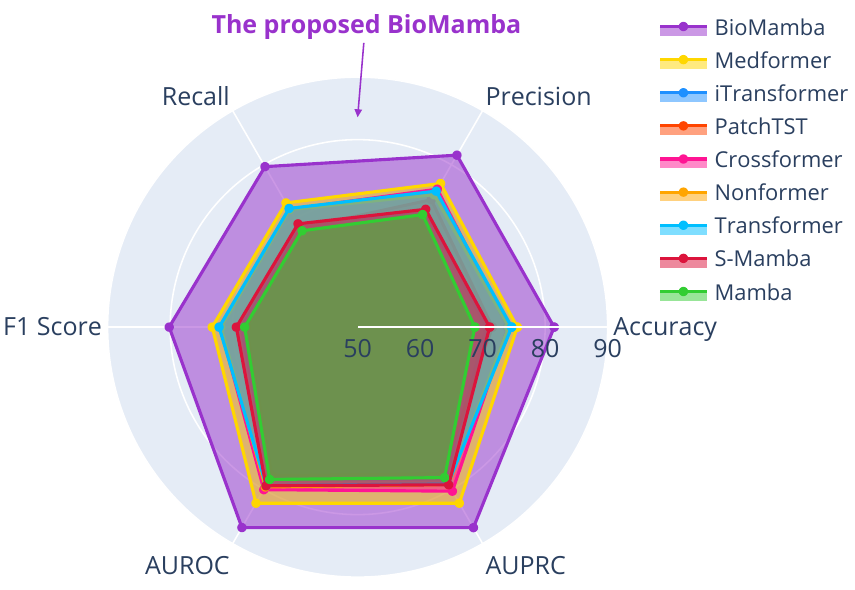}
\caption{Our BioMamba consistently outperforms state-of-the-art biosignals classification methods across six quality evaluation metrics with the average six datasets results.}
\label{fig: radar_average_results}
\end{figure}

\begin{figure}[t]
    \centering
    \includegraphics[width=1\columnwidth]{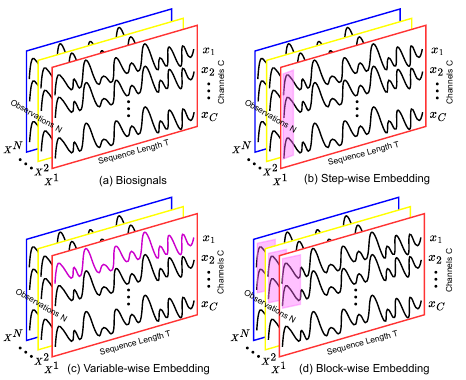} 
    \caption{The biosignals dimension information and three main types of embeddings.
}
    \label{fig: embedding_overview}
\end{figure}

Biosignals are physiological electrical information from the human body, measured as physical quantities through specialized sensors~\cite{hinrichs2020comparison, zhao2021ultra, xu2023ear}. These signals play a crucial role in various medical fields. For example, electroencephalograms (EEGs), which record neural electrical activity via scalp-mounted sensors, are routinely utilized in diagnosing seizure disorders. Similarly, electrocardiograms (ECGs), which capture the heart's electrical activity through surface electrodes, are indispensable for assessing cardiac arrhythmias and other pathologies affecting heart muscle function.
With advancements in wearable technology~\cite{tan2017locus, iqbal2021advances, goh2024walkingwizard}, access to such data has become significantly more feasible. In this paper, we aim to explore a novel framework to enhance the effective utilization of biosignal information for improved human health and well-being.

A variety of deep learning methods has been advanced for effectively modeling time-series information, including biosignals. Transformer-based methods, in particular, have shown outstanding performance in analyzing time series across various applications such as forecasting~\cite{zhang2023crossformer, liu2023itransformer}, generation~\cite{coletta2024constrained, qian2024timeldm}, and disease detection~\cite{wang2024medformer, mohammadi2024eeg2rep}.
For instance,
Medformer~\cite{wang2024medformer} presents a multi-granularity patching transformer adapted for medical time-series classification.
EEG2Rep~\cite{mohammadi2024eeg2rep} introduces an innovative self-supervised approach to tackle the inherent challenges of learning EEG data representations.
iTransformer~\cite{liu2023itransformer} uses the variable-wise embedding and maps the entire variable into a temporal token, which successfully reasons about interrelationships between variables.
PatchTST~\cite{nie2022time} segments time-series by dividing the sequence into patches, allowing for an increased input length while reducing redundant information.
Recently, Mamba-based methods have shown impressive capability in time-series analysis. As an example, S-Mamba~\cite{wang2024mamba} achieves leading-edge performance in time-series forecasting while requiring significantly lower computational overhead compared to Attention-based methods.

However, despite strong empirical performance when applied to biosignals, existing Attention-based and Mamba-based methods still fall short in practical applications.
We detail the issues from four aspects. 
\scalebox{1.1}{\ding{172}}. Attention-based methods face challenges with \textbf{inefficient learning and high computational overhead}, the quadratic complexity has led to substantial GPU memory and FLOPs, making them unsuitable for edge applications ( see Table~\ref{tab: complexity} ).
\scalebox{1.1}{\ding{173}}. Although Mamba-based methods perform well in general time-series analysis, they face challenges with biosignals, such as EEG data. Biosignals present unique characteristics—including high noise levels, non-stationarity, and complex temporal dependencies—which differ substantially from other types of time-series information, often resulting in \textbf{suboptimal performance}.
\scalebox{1.1}{\ding{174}}. Most existing approaches focus only on time-domain embeddings, \textbf{overlooking the benefits of frequency-domain information} (see Figure~\ref{fig: crowdsource_dataset}). The frequency domain captures essential periodic patterns, improves robustness to noise, and enables multi-scale feature extraction, which is crucial for accurately interpreting complex biosignals.
\scalebox{1.1}{\ding{175}}. A widely adopted approach is to apply dense FFN to extract non-linear transformations in latent representations. However, MLP-based FFNs commonly face \textbf{limitations in efficiency and generalization}, as they often handle redundant information and are prone to overfitting when trained on limited datasets, undermining training effectiveness.

\begin{figure*}[t]
    \centering
    \includegraphics[width=0.98\linewidth]{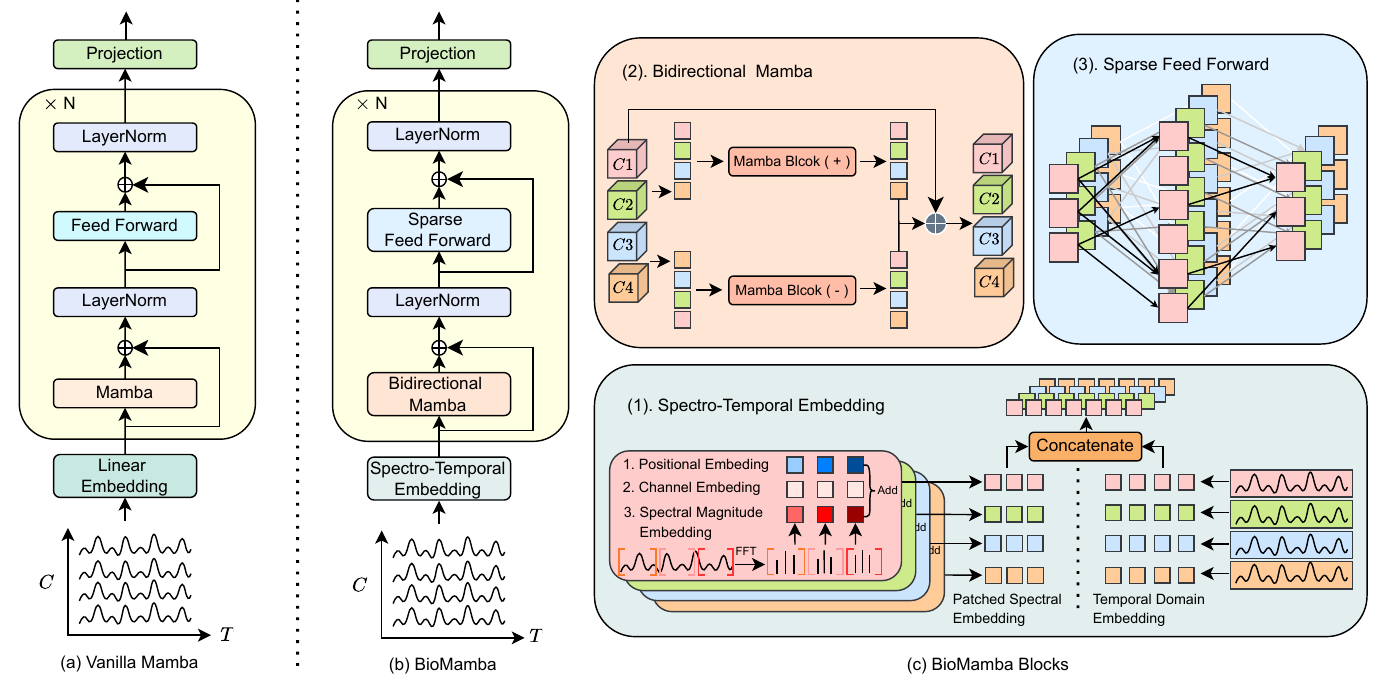} 
    \caption{An overview of the proposed BioMamba. (a)-(b) Comparison between the  Blocks of vanilla Mamaba and the proposed BioMamba. (c) Details of BioMamba blocks: (1). Spectro-Temporal Embedding strategy. (2). Bidirectional Mamba framework. (3). Sparse Feed Forward layers.
}
    \label{fig: method_overview}
\end{figure*}

In this paper, to improve learning efficiency and address the issues of existing work, an innovative biosignal classification method is introduced, \textbf{BioMamba}, where we utilize \textbf{Spectro-Temporal Embedding} for the \textbf{Bidirectional Mamba} blocks with the \textbf{Sparse Feed Forward} policy. The overall pipeline in Figure~\ref{fig: method_overview}, our approach introduces three key components to address these challenges. As can be seen, BioMamba employs a Spectro-Temporal Embedding technique that concatenates frequency-domain and time-domain information, allowing it to capture long-term dependencies by leveraging both spectral and temporal features. 
BioMamba engages in a bidirectional scanning approach, which processes embedding from both forward and backward perspectives. This enables the model to capture comprehensive contextual information across sequences and enrich feature representation with linear complexity.
The Sparsity Feed Forward module in BioMamba preserves only within the Subset Weights, enhancing both computational efficiency and model generalization.
\textbf{Specifically, the Spectro-Temporal Embedding is employed to tackle issues \scalebox{1.1}{\ding{173}} and \scalebox{1.1}{\ding{174}}, the Bidirectional Mamba block addresses issue \scalebox{1.1}{\ding{172}}, and the Sparsity Feed Forward resolves issue \scalebox{1.1}{\ding{175}}}.

We conduct an in-depth validation of the performance and efficiency of our proposed approach against eight baselines across six datasets. 
The results demonstrate that BioMamba achieves new state-of-the-art performance on five out of six datasets (see Table~\ref{tab: overall_performance}). 
The main contributions of BioMamba are as follows:
\begin{itemize}
    \item \textbf{Reliability.} We introduce a pioneering biosignal analysis architecture called {BioMamba}. This architecture employs a Spectro-Temporal Embedding strategy for biosignal token extraction, which integrates both frequency-based characteristics and temporal patterns. BioMamba consistently achieves improvements in performance for biosignal classification across six evaluation metrics.
    \item  \textbf{Efficiency.} We propose a Bidirectional Mamba framework with Sparse Feed Forward layers to enable effective learning of biosignal sequences compared to existing approaches.
    \item \textbf{Generality.} Evaluated on a diverse set of biosignal classification benchmarks and compared with strong baselines, including Attention-based and Mamba-based architectures, our model achieves new state-of-the-art performance on most tasks. It demonstrates strong adaptability and effectiveness across a wide range of domains and applications.
\end{itemize}


\section{Related Works}
\label{sec: related_works}

\textbf{Biosignals Classification.} Biosignals represent time-series data collected from human biological systems, encompassing EEG~\cite{tang2021self, qu2020ensemble}, ECG~\cite{xiao2023deep, wang2023hierarchical}, EMG~\cite{xiong2021deep, dai2022mseva}, EOG~\cite{jiao2020driver}, and other types~\cite{imtiaz2021systematic}. These signals are pivotal in applications such as disease diagnosis~\cite{liu2021deep}, emotion recognition~\cite{li2022eeg}, and fitness tracking~\cite{mun2024assessment}. The goal of biosignal classification is to predict categorical labels from these time-series inputs, facilitating tasks like Parkinson's disease detection~\cite{aljalal2022detection}, Alzheimer's disease classification~\cite{vicchietti2023computational}, and myocardial infarction identification~\cite{al2023machine}. 
Recent approaches for biosignal classification often rely on deep-learning models with CNNs, GNNs, and Transformers. For example, EEGNet~\cite{lawhern2018eegnet}, EEG-Conformer~\cite{song2022eeg}, Medformer~\cite{wang2024medformer}, and REST~\cite{afzal2024rest} have shown strong performance across various biosignal classification tasks.


\textbf{State Space Models.} 
Although variants of Attention-based models have achieved remarkable performance in sequence classification capability. The quadratic complexity concerning sequence length makes it computationally expensive and memory-intensive for long sequences, which limits the scalability of Attention-based methods in applications requiring extended sequences, such as speech and biosignals. To overcome the limitations of Attention-based methods, State Space Models (SSMs) have been integrated with deep 
learning to address the problem of long-range dependencies. 
Multiple optimized SSM variants, including S4~\cite{gu2021efficiently}, H3~\cite{fu2022hungry}, S5~\cite{smith2022simplified}, and Gated State Space~\cite{mehta2022long}, have been introduced to elevate both performance and efficiency in practical applications.
Recently, Mamba~\cite{gu2023mamba} has been proposed, surpassing previous methods by implementing a data-driven selection mechanism based on S4~\cite{gu2021efficiently}. This mechanism efficiently chooses important information from input sequence elements and captures long-range dependencies that scale with sequence length. With its linear learning complexity in handling long sequences, Mamba has seen broad adoption across various domains, including computer vision~\cite{zhu2024vision, shi2024multi} and natural language processing~\cite{pioro2024moe, he2024densemamba}.

\textbf{Time-Series Embedding.}
By acting as space transformations $\mathbb{R}^T \mapsto \mathbb{R}^E$, embedding methods facilitate the conversion of discrete and sparse features into continuous and dense vectors, laying a robust groundwork for success in multiple areas of machine learning~\cite{vaswani2017attention, dosovitskiy2020image}. In time-series analysis frameworks, existing embedding methods can be categorized into three main types (see Figure~\ref{fig: embedding_overview}):
(1) Step-wise Embedding: This approach considers each time step individually, embedding it into the unified token space. The Transformer~\cite{vaswani2017attention} exemplifies this by using a single cross-channel timestamp as the token for each time step.
(2) Variable-wise Embedding: This method treats each variable independently, embedding them separately before combining. iTransformer~\cite{liu2023itransformer} follows this approach, embedding each variable on its own and mapping the entire variable set into the time-wise tokens.
(3) Block-wise Embedding: This approach divides the time-series into fixed-size blocks or patches, capturing local temporal patterns within each block. PatchTST~\cite{nie2022time} demonstrates this method by leveraging patch embeddings to enhance feature extraction across segments of time. 


\section{Methodology}
\label{sec: method}

In this section, we formally introduce \textbf{BioMamba} (\textbf{Bio}signals Classification with Bidirectional \textbf{Mamba}). Figure~\ref{fig: method_overview} illustrates the overall architecture of BioMamba along with the details of its core blocks. We first formulate the biosignal classification task. Then, we introduce the pipeline of proposed BioMamba. Finally, we provide a detailed explanation of each BioMamba block.

\subsection{Preliminaries}
\label{sec:preliminaries}

\textbf{Problem Statement.} As the Figure~\ref{fig: embedding_overview} (1), given a multivariate biosignal dataset with corresponding labels $(\mathbf{X}, \mathbf{Y})$, where $\mathbf{X} = \left[\mathbf{x}^1, \mathbf{x}^2, \ldots, \mathbf{x}^N\right]$ and $\mathbf{Y} = \left[\mathbf{y}^1, \mathbf{y}^2, \ldots, \mathbf{y}^N\right]$, each multivariate time-series $\mathbf{x}$ has the form $\mathbf{x} = \left(\mathrm{x}_1, \ldots, \mathrm{x}_C\right) \in \mathbb{R}^{T \times C}$. Here, $N$ is the number of observations, $T$ denotes the sequence length, and $C$ is the number of channels. The objective of BioMamba is to learn a classifier $f_\theta$ that maps each series $\mathbf{x}^n$ to its corresponding class within ${1, \ldots, K}$, where $K$ is the total number of classes.

\textbf{State Space Models.}
Originating from the Kalman filter~\cite{kalman1960new}, SSMs can be regarded as linear time-invariant  (LTI) systems that map the input stimulation $x(t) \in \mathbb{R}$ to response $y(t) \in \mathbb{R}$ through the hidden state ${h}(t) \in \mathbb{R}^M$. 
Specifically, continuous-time SSMs can be formulated as linear ordinary differential equations (ODEs) as follows:
\begin{equation}
\begin{aligned} h^{\prime}(t) & =\boldsymbol{A} h(t)+\boldsymbol{B} x(t)
\\ y(t) & =\boldsymbol{C} h(t) 
\end{aligned}
\label{eq: ssm1}
\end{equation}
where $h^{\prime}(t)=\frac{d h(t)}{d t}$, and $\mathbf{A} \in \mathbb{R}^{M \times M}, \mathbf{B} \in \mathbb{R}^{M \times 1}, and~\mathbf{C} \in \mathbb{R}^{1 \times M}$ are learnable matrices of the SSMs.
Then, the continuous sequence is discretized by a step size $\Delta$, and the discretized SSM model is represented as: 
\begin{equation}
\begin{aligned} h_t & =\overline{\boldsymbol{A}} h_{t-1}+\overline{\boldsymbol{B}} x_t \\ y_t & =\boldsymbol{C} h_t 
\end{aligned}
\label{eq: ssm2}
\end{equation}
where $h_t$ and $x_t$ are the state vector and input vector at time $t$, respectively, and  $\overline{\boldsymbol{A}}=\exp (\Delta \boldsymbol{A})$ and $\overline{\boldsymbol{B}}=(\Delta \boldsymbol{A})^{-1}(\exp (\Delta \boldsymbol{A})-I) \cdot \Delta \boldsymbol{B}$. Since transitioning from continuous form $(\Delta, \boldsymbol{A}, \boldsymbol{B}, \boldsymbol{C})$ to discrete form $(\overline{\boldsymbol{A}}, \overline{\boldsymbol{B}}, \boldsymbol{C})$, the model can be efficiently calculated using a linear recursive approach.

To further accelerate computation,~\cite{gu2021efficiently} expanded the SSM computation into a convolution with a structured convolutional kernel $\boldsymbol{K} \in \mathbb{R}^L$ :
\begin{equation}
\begin{aligned} 
\bar{\boldsymbol{K}} &\triangleq\left(\boldsymbol{C} \bar{\boldsymbol{B}}, \boldsymbol{C} \overline{\boldsymbol{A} \boldsymbol{B}}, \cdots, \boldsymbol{C} \bar{\boldsymbol{A}}^{L-1} \bar{\boldsymbol{B}}\right) \\
y&=x * \bar{\boldsymbol{K}}
\end{aligned}
\label{eq: ssm3}
\end{equation}
where $L$ is the length of the input sequence and $*$ denotes the convolution operation. Based on the mentioned discrete State-Space Equations~\ref{eq: ssm2}, Mamba~\cite{gu2023mamba} introduces data dependency into the model parameters, enabling the model to selectively propagate or forget information based on the sequential input tokens. In addition, it utilizes a parallel scanning algorithm to accelerate the equation-solving process, making it highly compatible with hardware implementations.

\subsection{Overall Architecture}
\label{sec:architecture}
In this paper, we propose BioMamba, a biosignal classification method designed to overcome the inefficiencies and performance limitations of existing approaches. As shown in Figure~\ref{fig: method_overview}, our BioMamba mainly consists of three key modules: the Spectro-Temporal Embedding, the Bidirectional Mamba, and the Sparse Feed Forward. Each serves a specific purpose in the overall pipeline, which is to tackle the limitations of existing methods. Figure~\ref{fig: method_overview} (c) illustrates the details of three components.
This procedure can be described as algorithm~\ref{alg: biomamba}.
In the following sections, we provide comprehensive explanations and illustrations for each of these components.

\begin{algorithm}[htpb]
   \caption{The BioMamba Algorithm}
   \label{alg: biomamba}
\begin{algorithmic}
\small
   \STATE \textbf{Input:} $\mathbf{X} = \left[\mathbf{x}^1, \mathbf{x}^2, \ldots, \mathbf{x}^B\right]:(B, T, C)$
   \STATE \textbf{Output:} $\hat{\mathbf{Y}} = \left[\hat{\mathbf{y}}^1, \hat{\mathbf{y}}^2, \ldots, \hat{\mathbf{y}}^B\right]:(B,K)$
   \STATE $\mathbf{X}:(B, C, T) \leftarrow \operatorname{Transpose}(\mathbf{X})$
   \STATE $\mathbf{Z}:(B, E, D) \leftarrow \operatorname{Spetctro-Temporal~ Embedding}(\mathbf{X})$
   \FOR{$m$ \textbf{in layers}}
    \STATE $\operatorname{Bidrirectional~Mamba~:}$ 
    \STATE \quad $\mathbf{Z^{m-1}_1}:(B, E, D) \leftarrow \operatorname{Mamba(+)}(\mathbf{Z^{m-1}})$
    \STATE \quad $\mathbf{Z^{m-1}_2}:(B, E, D) \leftarrow \operatorname{Re}(\operatorname{Mamba(-)}(\operatorname{Re}(\mathbf{Z^{m-1}})))$
    \STATE \quad \quad \quad \quad  $ /* where~Re~is~the~Reverse */$ 
   \STATE $\mathbf{Z^{m-1}}:(B, E, D) \leftarrow \operatorname{LN}((\mathbf{Z^{m-1}_1} + \mathbf{Z^{m-1}_2}) + \mathbf{Z^{m-1}})$
   \STATE $\mathbf{Z^{m-1}_s}:(B, E, D) \leftarrow \operatorname{Sparse~Feed~Forward}(\mathbf{Z^{m-1}})$
  \STATE $\mathbf{Z^{m}}:(B, E, D) \leftarrow \operatorname{LN}(\mathbf{Z^{m-1}_s}+ \mathbf{Z^{m-1}})$
   \ENDFOR
   \STATE $\hat{\mathbf{Y}}:(B,K) \leftarrow \operatorname{Projection}(\mathbf{Z^m})$
\end{algorithmic}
\end{algorithm}

\subsection{Spectro-Temporal Embedding}
\label{sec: embedding}
As shown in Figure~\ref{fig: method_overview} (1),
We propose Spectro-Temporal Embedding (STE), a fusion embedding strategy that captures both frequency-based features and temporal patterns to achieve a richer representation of the input biosignals.
Specifically, consider the input $\mathbf{x} = \left(\mathrm{x}_1, \ldots, \mathrm{x}_C\right) \in \mathbb{R}^{T \times C}$. The Spectro-Temporal Embedding consists of two types: the Patched Spectral Embedding (PSE) and the Temporal Domain Embedding (TDE).

For the Patched Spectral Embedding, we apply a segmentation approach for the frequency domain with a defined frequency resolution to obtain segmented biosignals $\mathbf{x_{seg}} = \left( [\mathrm{x}_1, \ldots, \mathrm{x}_{c_0}], [\mathrm{x}_{c_1}, \ldots,],[\ldots] \right) $. Then, we adopt Fast Fourier Transform (FFT)~\cite{Nussbaumer1982} to extract spectral information $ \mathbf{ FFT } \left( \mathbf{x_{seg}}\right) $ for each samples.
After that, we utilize a fully connected network to learn the Spectral Magnitude Embedding $ \mathbf{ FC } \left( \mathbf{ FFT } \left( \mathbf{x_{seg}}\right) \right) $.
We learn Channel Embedding (CE) from all the channels $C$ and add to the corresponding Spectral Magnitude Embedding. Meanwhile, within the channel, we adopt Positional Embedding (PE) for the Spectral Magnitude Embedding.
So the Patched Spectral Embedding can be listed as follows:
\begin{equation}
\resizebox{0.9\columnwidth}{!}{$
\mathbf{PSE}  = \mathbf{PE} \left[ \mathbf{ FC } \left( \mathbf{ FFT } \left( \mathbf{x_{seg}}\right) \right) + \mathbf{CE}\right] + \mathbf{ FC } \left( \mathbf{ FFT } \left( \mathbf{x_{seg}}\right) \right) + \mathbf{CE}$}
\label{eq:sme}
\end{equation}
And the $\mathbf{PSE} \in  \mathbb{R}^{S \times D}$, where $S $ is the sample amount for all samples and $D$ is the hidden dimension of BioMamba.

For the Temporal Domain Embedding, we employ the variable-wise embedding strategy, given the input $\mathbf{x}$, the temporal-based features can be listed as follows: 
\begin{equation}
\resizebox{0.8\columnwidth}{!}{$
\mathbf{TDE} = \left(\mathbf{FC}(\mathrm{x}_1), \ldots, \mathbf{FC}(\mathrm{x}_C)\right) \in \mathbb{R}^{C \times D}$}
\label{eq:tde}
\end{equation}
where the $D$ is the hidden dimension.
Finally, the Spectro-Temporal Embedding concatenates the Patched Spectral Embedding with the Temporal Domain Embedding in a hidden dimension.
\begin{equation}
\resizebox{0.8\columnwidth}{!}{$
\mathbf{STE} = \mathbf{Concat} \left(\mathbf{PSE},\mathbf{TDE} \right) \in \mathbb{R}^{E\times D} $}
\label{eq:tde}
\end{equation}
where $E= C + S$ is the combined dimension.
Based on the results in Table~\ref{tab: ablation_of_embedding}, Patched Spectral Embedding significantly enhances BioMamba's ability to interpret complex biosignals by integrating spectral insights with time-based context.
We also provide the ablation study of frequency resolution in Table~\ref{tab: ablation_of_frequency_resolution} to evaluate the effect of frequency bins and window shifts. 

\subsection{Bidirectional Mamba}
\label{sec:bidirectional_mamba}
Despite the unidirectional scan in Mamba offering promising advantages for modeling causal sequential data, it lacks the ability to capture global inter-variate mutual information~\cite{wang2024mamba,zhu2024vision}. However, for modeling biosignals, which often have complex global dependencies and local interactions. To address this, We capitalize on the advantages of the bidirectional structure to devise vanilla mamba blocks, enabling the modeling of sequence information in both forward and reverse spectro-temporal directions. As shown in Figure~\ref{fig: method_overview} (2), given the Spectro-Temporal Embedding tokens $Z \in \mathbb{R}^{E \times D} $, we utilize two Mamba blocks to construct a bidirectional architecture and define the representations as follows:
\begin{equation}
\begin{aligned} 
\boldsymbol{Z_1}& = \mathbf{Mamba(+)}(\boldsymbol{Z})  \in \mathbb{R}^{E \times D} \\ 
\boldsymbol{Z_2}& = \mathbf{Reverse}\mathbf{(Mamba(-)}(\mathbf{Reverse}(\boldsymbol{Z}))) \in \mathbb{R}^{E \times D}
\end{aligned}
\label{eq: bimamba}
\end{equation}
Following this, we incorporate a fusion tactic and a residual connection to generate the results of the Bidirectional Mambablock.

\begin{equation}
\begin{aligned} 
\boldsymbol{Z'} &= \boldsymbol{Z_1} +  \boldsymbol{Z_2} \in \mathbb{R}^{E \times D} \\
\boldsymbol{Z^{''}} &= \boldsymbol{Z'} +\boldsymbol{Z}\in \mathbb{R}^{E \times D}
\end{aligned}
\label{eq: bimamba_s}
\end{equation}

\subsection{Sparse Feed Forward}
\label{sec: sparse_feed_forward}
The standard Attention-based or Mamba-based methods for time-series analysis regularly incorporate dense FFN for non-linear transformations within latent spaces. However, the FFN requires substantial computational resources and accounts for about two-thirds of a Transformer layer’s
parameters~\cite{geva2020transformer}, which can make these models prone to overfitting, especially when the dataset is small.
In this paper, we embrace the Sparse Feed Forward layer to enhance feature extraction capabilities, with the goal of achieving high computational efficiency in biosignal analysis. 

See Figure~\ref{fig: method_overview} (3), we take on a random sampling policy to optimize the weights $w$ of the Feed Forward layer. Specifically, we apply a subset $S$ randomly selected from the dense weight indices. The subset $S$ specifies which weights remain active, allowing control over the fraction of weights retained. 
\begin{equation}
w_i= \begin{cases}w_i, & \text { if } i \in S \\ 0, & \text { if } i \notin S\end{cases}
\end{equation}
where the subset $S$ is defined in Set~\ref{eq:subset_s}, and $R$ is computed from Equation~\ref{eq: s}. The Sparsity is a tunable hyperparameter, We evaluate different Sparsity settings in the ablation study in Table~\ref{tab: ablation_of_sparsity}.
\begin{equation}
\resizebox{0.5\textwidth}{!}{%
$S = \{ i \mid i \in \operatorname{Subset}(\{0, 1,  \dots, \text{In\_Features} \times \text{Out\_Features} - 1\}, \mathrm{R}) \}$
}
\label{eq:subset_s}
\end{equation}
\begin{equation}
\resizebox{0.4\textwidth}{!}{%
$\mathcal{R} = \boldsymbol{Round} \left[ (1 - \text{Sparsity}) \times \text{In\_Features} \times \text{Out\_Features} \right ]$%
}
\label{eq: s}
\end{equation}


\begin{table*}[htpb]
\centering
\resizebox{\textwidth}{!}{%
\begin{tabular}{ccccccccccccc}
\toprule
Datasets & Subject & Sample & Class &  Channel  & Timestamps & Sampling Rate & Modality &  File Size & Tasks \\
\hline 
APAVA & 23 & 5,967 & 2 & 16 & 256 & 256 Hz & EEG & 186 MB &  Alzheimer’s disease Classification  \\
TDBrain & 72 & 6,240 & 2 & 33 & 256 & 256 Hz & EEG & 571 MB  & Parkinson’s disease Detection  \\
Crowdsourced & 13 & 12,296 & 2  & 14 &   256  & 128 Hz  & EEG  & 620 MB & Eyes open/close Detection  \\
STEW  & 48 & 26,136 & 2 & 14 & 256 &   128 Hz  &   EEG &  682 MB & Mental workload Classification  \\
DREAMER  & 23 & 77,910 & 2  & 14 &   256  &  128 Hz & EEG & 2.00 GB & Emotion Detection \\
PTB & 198 & 64,356 & 2 & 15 & 300 & 250 Hz & ECG & 2.15 GB & Myocardial Infarction  \\
\bottomrule
\end{tabular}}
\caption{Overview of biosignal datasets}
\label{tab: all_datasets}
\end{table*}

\section{Experiments}
\label{sec: exp}

In this section, we present extensive experiments to demonstrate the advantages of our proposed method, BioMamba, focusing on both classification performance and computational efficiency. To achieve this, we compare BioMamba with eight baseline models, covering a diverse range of approaches, including both Attention-based and Mamba-based architectures. 
The datasets selection span six diverse tasks for binary clinical diagnosis. Additionally, we present further experiments in multiclass classification. All of these experiments comprehensively demonstrate the applications of Biomamba in biosignals, providing a new baseline for real-world applications.
We used two NVIDIA RTX 4090 24GB GPUs with an Intel(R) Xeon(R) Gold 6230 CPU @ 2.10GHz for all experiments of our BioMamba and eight baseline models.

\subsection{Setups}

\textbf{Datasets.} We conduct a thorough experimental analysis on six datasets, including five EEG datasets and one ECG dataset: APAVA~\cite{escudero2006analysis}, TDBrain~\cite{van2022two}, Crowdsourced~\cite{williams2023crowdsourced}, STEW~\cite{lim2018stew}, DREAMER~\cite{katsigiannis2017dreamer}, and PTB~\cite{goldberger2000physiobank}. An overview of the datasets is available in Table~\ref{tab: all_datasets}, and we also present the eyes closed and open states in both the frequency and time domains in Figure~\ref{fig: crowdsource_dataset}. The additional descriptions, including details on data preprocessing, can be found in Appendix~\ref{sec: datasets}.

\textbf{Baselines.} We compare against eight state-of-the-art time-series methods. The first two are Mamba-based models: Mamba~\cite{gu2023mamba} and S-Mamba~\cite{wang2024mamba}. We also evaluate the vanilla Transformer~\cite{vaswani2017attention} for biosignal classification. Additionally, we assess six recent pioneering methods, including Nonformer~\cite{liu2022non}, Crossformer~\cite{zhang2023crossformer}, PatchTST~\cite{nie2022time}, iTransformer~\cite{liu2023itransformer}, and Medformer~\cite{wang2024medformer}. Notably, the original Mamba~\cite{gu2023mamba} and S-Mamba~\cite{wang2024mamba} methods do not provide a detailed evaluation for biosignals; our paper addresses this gap by offering a standardized performance evaluation of Vanilla Mamba and S-Mamba. More details of the baselines are listed in Appendix~\ref{sec: Baselines}

\textbf{Implementation.} We use six evaluation metrics: Accuracy, Macro-averaged Precision, Macro-averaged Recall, Macro-averaged F1 score, Macro-averaged AUROC, and Macro-averaged AUPRC. Each dataset is partitioned into subject-wise train, validation, and test sets, simulating real-world biosignal-based disease diagnosis scenarios and challenging models to capture generalized patterns. Training is performed with five random seeds on these fixed sets, allowing us to compute the mean and standard deviation of the model performances. Additional implementation details are provided in Appendix~\ref{sec: implementation_details}.

\begin{table*}[t!]
\centering
\resizebox{0.86\textwidth}{!}{%
\begin{tabular}{clcccccc|cc}
\toprule
Datasets & Models & Accuracy & Precision & Recall & F1 score & AUROC & AUPRC & Params (M) & FLOPs (G)  \\
\hline 
\multirow{11}{*}{\begin{tabular}{l}
APAVA \\
\end{tabular}} 
& Mamba & $75.75 _{\pm 1.51}$ & $76.08 _{\pm 1.43}$ & $73.05 _{\pm 1.89}$ & $73.64 _{\pm 1.93}$ & $85.94 _{\pm 1.29}$ & $84.27 _{\pm 1.37}$ & $0.75$M & $0.41$G \\
& S-Mamba & $76.59 _{\pm 1.61}$ & $77.19 _{\pm 1.29}$ & $74.00 _{\pm 2.51}$ & $74.53 _{\pm 2.45}$ & $86.36 _{\pm 1.17}$ & $84.88 _{\pm 1.25}$ & $1.07$M & $0.58$G \\
& Transformer & $75.61 _{\pm 6.22}$ & $77.41 _{\pm 7.89}$ & $ 72.04_{\pm 6.51}$ & $72.66 _{\pm 7.06}$ & $69.73 _{\pm 5.91}$ & $70.63 _{\pm 6.70}$ & $0.87$M & $9.78$G \\
& Nonformer & $69.36 _{\pm 7.50}$ & $68.99 _{\pm 8.02}$ & $ 67.62_{\pm 7.09}$ & $67.61 _{\pm 7.64}$ & $69.50 _{\pm 5.88}$ & $69.36 _{\pm 6.50}$ & $0.94$M & $9.79$G \\
& Crossformer & $73.78 _{\pm 2.78}$ & $79.18 _{\pm 3.43}$ & $ 68.89_{\pm 3.45}$ & $68.80 _{\pm 4.29}$ & $75.60 _{\pm 6.48}$ & $74.87 _{\pm 6.09}$ & $5.23$M & $6.72$G \\
& PatchTST & $67.11 _{\pm 2.65}$ & $78.68 _{\pm 1.18}$ & $ 60.04_{\pm 3.35}$ & $56.07 _{\pm 5.29}$ & $65.72 _{\pm 2.74}$ & $67.88 _{\pm 2.22}$ & $0.93$M & $13.86$G \\
& iTransformer & $74.91 _{\pm 0.62}$ & $75.61 _{\pm 1.25}$ & $ 72.28_{\pm 1.90}$ & $72.64 _{\pm 1.78}$ & $85.85 _{\pm 1.12}$ & $84.22 _{\pm 1.34}$ & $0.83$M & $0.44$G \\
& Medformer & $77.81 _{\pm 2.67}$ & $80.68 _{\pm 4.03}$ & $74.27 _{\pm2.69 }$ & $75.09 _{\pm 2.89}$ & $81.05 _{\pm 4.62}$ & $81.59 _{\pm 4.29}$ & $7.41$M & $21.30$G \\
\cmidrule(lr){2-10} 
& BioMamba (Ours) & \deepred{$84.95 _{\pm 1.35}$} & \deepred{$85.72 _{\pm 1.95}$} & \deepred{$83.15 _{\pm 1.13}$} & \deepred{$83.95 _{\pm 1.32}$} & \deepred{$93.79 _{\pm 1.39}$} & \deepred{$93.52 _{\pm 1.40}$} & \deepreds{$0.97$M} & \deepreds{$1.61$G} \\
& Improve. & \deepred{$ +7.14 $} & \deepred{$ +5.04 $} & \deepred{$ +8.88 $} & \deepred{$ +8.86 $} & \deepred{$ +12.74$} & \deepred{$ +11.93 $} & \deepred{$ 8\times $} & \deepred{$ 13\times $} \\
\hline 
\multirow{11}{*}{\begin{tabular}{l}
TDBrain \\
\end{tabular}} 
& Mamba & $72.52 _{\pm 0.64}$ & $72.67 _{\pm 0.69}$ & $72.52 _{\pm 0.64}$ & $72.48 _{\pm 0.63}$ & $80.88 _{\pm 1.08}$ & $80.68_{\pm 1.05}$ & $0.76$M & $0.81$G\\
& S-Mamba & $73.40 _{\pm 0.97}$ & $73.55_{\pm 1.02}$ & $73.40 _{\pm 0.97}$ & $73.35 _{\pm 0.96}$ & $81.51 _{\pm 1.17}$ & $81.20 _{\pm 1.25}$ & $1.07$M & $1.15$G\\
& Transformer & $87.88_{\pm 3.35}$ & $88.84 _{\pm 2.37}$ & $ 87.88_{\pm 3.35}$ & $87.77 _{\pm 3.50}$ & $96.50 _{\pm0.93 }$ & $96.29 _{\pm 1.27}$ & $0.87$M& $9.84$G\\
& Nonformer & $86.18_{\pm 2.51}$ & $87.32_{\pm 2.05}$ & $ 86.19_{\pm 2.51}$ & $86.07 _{\pm 2.57}$ & $96.19_{\pm 1.16}$ & $96.26 _{\pm 1.11}$ & $0.96$ M & $9.84$ G \\
& Crossformer & $82.79 _{\pm 1.99}$ & $83.13 _{\pm 2.04}$ & $ 82.79_{\pm 1.99}$ & $82.75 _{\pm 1.99}$ & $92.06 _{\pm 1.98}$ & $92.19 _{\pm 2.10}$ & $5.29$ M & $13.75$ G \\
& PatchTST & $73.33 _{\pm 2.82}$ & $73.45 _{\pm 2.79}$ & $ 73.33_{\pm 2.82}$ & $73.30 _{\pm 2.84}$ & $80.52 _{\pm 4.53}$ & $78.12 _{\pm 5.46}$ & $1.07$ M & $28.59$ G \\
& iTransformer & $74.77_{\pm 0.57}$ & $74.97_{\pm 0.55}$ & $ 74.77_{\pm 0.59}$ & $74.72 _{\pm 0.61}$ & $83.36_{\pm 1.00}$ & $83.52 _{\pm 0.93}$ & $0.84$ M & $0.93$ G \\
& Medformer & $88.31 _{\pm 1.65}$ & $88.43_{\pm 1.51}$ & $88.31 _{\pm1.65 }$ & $88.30 _{\pm1.66}$ & $95.90 _{\pm 0.72}$ & $96.00 _{\pm 0.64}$ & $3.52$M& $4.31$G\\
\cmidrule(lr){2-10} 
& BioMamba (Ours) & \deepred{$96.77 _{\pm 1.94}$} & \deepred{$96.90 _{\pm 1.71}$} & \deepred{$96.77 _{\pm 1.94}$} & \deepred{$96.77 _{\pm 1.95}$} & \deepred{$99.44 _{\pm 0.49}$} & \deepred{$99.42 _{\pm 0.51}$} & \deepreds{$0.83$M} & \deepreds{$2.22$G} \\
& Improve. & \deepred{$ +8.46 $} & \deepred{$ +8.47 $} & \deepred{$ +8.46 $} & \deepred{$ +8.47 $} & \deepred{$ +3.54$} & \deepred{$3.42 $} & \deepred{$4\times $} & \deepred{$2\times $} \\
\hline
\multirow{11}{*}{\begin{tabular}{l}
Crowdsourced \\
\end{tabular}} 
& Mamba & $76.87 _{\pm 1.08}$ & $79.14 _{\pm 0.61}$ & $76.87 _{\pm 1.08}$ & $76.40 _{\pm 1.22}$ & $89.52 _{\pm 0.18}$ & $89.45_{\pm 0.30}$ & $0.75$M & $0.36$G \\
& S-Mamba & $76.44 _{\pm 0.87}$ & $78.51 _{\pm 0.50}$ & $76.43 _{\pm 0.87}$ & $76.00 _{\pm 1.00}$ & $89.01 _{\pm 0.90}$ & $89.05 _{\pm 0.96}$ & $1.07$M & $0.51$G \\
& Transformer & $80.13_{\pm 1.55}$ & $80.37 _{\pm 1.45}$ & $ 80.12_{\pm 1.55}$ & $80.08 _{\pm 1.57}$ & $88.61 _{\pm 1.49}$ & $88.07 _{\pm 1.84}$ & $0.87$M & $9.78$G \\
& Nonformer & $80.69 _{\pm 1.29}$ & $81.34_{\pm 1.55}$ & $ 80.69_{\pm 1.29}$ & $80.59_{\pm 1.29}$ & $88.81 _{\pm 1.17}$ & $87.86 _{\pm 1.24}$ & $0.94$ M & $9.78$ G \\
& Crossformer & $77.27 _{\pm 1.50}$ & $79.58 _{\pm 0.98}$ & $ 77.27_{\pm 1.50}$ & $76.81_{\pm 1.68}$ & $89.62 _{\pm 0.80}$ & $89.60 _{\pm 0.70}$ & $5.23$ M & $5.89$ G \\
& PatchTST & $85.83 _{\pm 1.95}$ & $86.29 _{\pm 2.03}$ & $ 85.83_{\pm 1.95}$ & $85.79_{\pm 1.95}$ & $93.73_{\pm 2.37}$ & $93.28 _{\pm 3.04}$ & $0.91$ M & $12.13$ G \\
& iTransformer & $73.71 _{\pm 3.31}$ & $76.79 _{\pm 2.46}$ & $ 73.71_{\pm 3.31}$ & $72.88_{\pm 3.76}$ & $86.83 _{\pm 1.73}$ & $86.69_{\pm 1.95}$ & $0.83$ M & $0.38$ G \\
& Medformer & $81.38 _{\pm 1.77}$ & $82.52_{\pm 1.31}$ & $81.38 _{\pm1.77 }$ & $81.21 _{\pm 1.89}$ & $91.58 _{\pm 0.89}$ & $91.52_{\pm 0.74}$ & $7.35$M & $21.27$G \\
\cmidrule(lr){2-10} 
& BioMamba (Ours) & \deepred{$89.84 _{\pm 0.72}$} & \deepred{$90.04 _{\pm 0.75}$} & \deepred{$89.83 _{\pm 0.72}$} & \deepred{$89.82 _{\pm 0.71}$} & \deepred{$96.88 _{\pm 0.34}$} & \deepred{$96.97 _{\pm 0.33}$} & \deepreds{$0.81$M} & \deepreds{$1.40$G} \\
& Improve. & \deepred{$ +8.46 $} & \deepred{$ +7.52$} & \deepred{$ +8.45 $} & \deepred{$ +8.61 $} & \deepred{$ +5.30 $} & \deepred{$ +5.45 $} & \deepred{$9\times $} & \deepred{$15\times $} \\
\hline \multirow{11}{*}{\begin{tabular}{l}
STEW \\
\end{tabular}} 
& Mamba & $63.42 _{\pm 1.77}$ & $64.15_{\pm 1.96}$ & $63.42 _{\pm 1.77}$ & $62.95 _{\pm 1.75}$ & $70.57 _{\pm 2.85}$ & $69.94_{\pm 2.99}$ & $0.75$M & $0.72$G \\
& S-Mamba & $67.65 _{\pm 0.61}$ & $68.28 _{\pm 0.64}$ & $67.65 _{\pm 0.61}$ & $67.36 _{\pm 0.61}$ & $76.01 _{\pm 0.47}$ & $75.38 _{\pm 0.44}$ & $1.07$M & $1.02$G \\
& Transformer & $77.20_{\pm 0.58}$ & $77.52 _{\pm 0.62}$ & $ 77.20_{\pm 0.58}$ & $77.14 _{\pm 0.58}$ & $84.70 _{\pm 0.64}$ & $83.92 _{\pm 0.72}$ & $0.87$M & $19.56$G \\
& Nonformer & $77.46 _{\pm 1.29}$ & $77.67 _{\pm 1.12}$ & $ 77.46_{\pm 1.29}$ & $77.41_{\pm 1.33}$ & $85.48 _{\pm 1.06}$ & $84.94 _{\pm 1.06}$ & $0.94$ M & $19.54$ G \\
& Crossformer & $76.78 _{\pm 0.75}$ & $77.13 _{\pm 0.71}$ & $ 76.78_{\pm 0.75}$ & $76.71_{\pm 0.77}$ & $84.89 _{\pm 0.83}$ & $84.36 _{\pm 0.84}$ & $5.23$ M & $11.78$ G \\
& PatchTST & $76.60_{\pm 1.24}$ & $76.84 _{\pm 1.02}$ & $ 76.60_{\pm 1.24}$ & $76.54_{\pm 1.29}$ & $85.51_{\pm 0.66}$ & $85.61_{\pm 0.56}$ & $0.91$ M & $24.26$ G \\
& iTransformer & $68.35 _{\pm 0.53}$ & $68.44 _{\pm 0.55}$ & $ 68.35_{\pm 0.53}$ & $68.31_{\pm 0.52}$ & $75.24 _{\pm 0.50}$ & $74.42 _{\pm 0.50}$ & $0.83$ M & $0.76$ G \\
& Medformer & $77.31 _{\pm 0.42}$ & $78.02_{\pm 0.87}$ & $77.31 _{\pm 0.42 }$ & $77.17 _{\pm 0.41}$ & $85.30 _{\pm 0.55}$ & $84.61_{\pm 0.38}$ & $7.35$M & $42.54$G \\
\cmidrule(lr){2-10} 
& BioMamba (Ours) & \deepred{$79.60 _{\pm 1.00}$} & \deepred{$79.65 _{\pm 1.03}$} & \deepred{$79.60 _{\pm 1.00}$} & \deepred{$79.59 _{\pm 0.99}$} & \deepred{$87.44 _{\pm 0.56}$} & \deepred{$87.27 _{\pm 0.53}$} & \deepreds{$0.73 $M} & \deepreds{$1.88$G} \\
& Improve. & \deepred{$ +2.29 $} & \deepred{$ + 1.63$} & \deepred{$ +2.29 $} & \deepred{$ +2.42 $} & \deepred{$ +2.14 $} & \deepred{$ +2.66 $} & \deepred{$10\times $} & \deepred{$23\times $} \\
\hline 
\multirow{11}{*}{\begin{tabular}{l}
DREAMER \\
\end{tabular}} 
& Mamba & $51.05 _{\pm 2.59}$ & $48.22_{\pm 2.97}$ & $48.33 _{\pm 2.85}$ & $48.17 _{\pm 2.95}$ & $50.82 _{\pm 3.76}$ & $52.08_{\pm 2.74}$ & $0.75$M & $1.43$G \\
& S-Mamba & $50.04 _{\pm 2.96}$ & $47.67 _{\pm 2.91}$ & $47.71 _{\pm 2.82}$ & $47.60 _{\pm 2.87}$ & \deepred{$50.86 _{\pm 2.43}$} & {$52.66 _{\pm 2.14}$} & $1.07$M & $2.03$G \\
& Transformer & $49.96_{\pm 2.87}$ & $46.85 _{\pm 2.89}$ & $ 47.01_{\pm 2.81}$ & $46.77_{\pm 2.84}$ & $46.02 _{\pm 1.18}$ & $48.68 _{\pm 0.54}$ & $0.87$M & $39.11$G \\
& Nonformer & $52.51_{\pm 1.35}$ & $48.83 _{\pm 1.66}$ & $ 48.99_{\pm 1.52}$ & $48.48_{\pm 1.78}$ & $47.53_{\pm 1.78}$ & $49.27_{\pm 1.71}$ & $0.94$ M & $39.12$ G \\
& Crossformer & $49.21_{\pm 2.90}$ & $46.85 _{\pm 2.34}$ & $ 46.93_{\pm 2.20}$ & $46.67_{\pm 2.29}$ & $46.00_{\pm 1.95}$ & $49.22 _{\pm 1.67}$ & $5.23$ M & $23.55$ G \\
& PatchTST & $48.88_{\pm 1.45}$ & $45.66 _{\pm 0.82}$ & $ 45.88_{\pm 0.90}$ & $45.60_{\pm 0.73}$ & $49.75_{\pm 2.04}$ & \deepred{$53.19 _{\pm 2.69}$} & $0.91$ M & $48.53$ G \\
& iTransformer & $48.89_{\pm 1.37}$ & $45.68 _{\pm 2.36}$ & $ 45.98_{\pm 2.16}$ & $45.68_{\pm 2.36}$ & $46.94_{\pm 2.12}$ & $48.98 _{\pm 1.82}$ & $0.83$ M & $1.52$ G \\
& Medformer & $50.52 _{\pm 1.64}$ & $48.19_{\pm 1.59}$ & $48.22 _{\pm 01.56 }$ & $48.16_{\pm 1.54}$ & $48.28 _{\pm 1.80}$ & $50.71_{\pm 2.25}$ & $7.35$M & $85.07$G \\
\cmidrule(lr){2-10} 
& BioMamba (Ours) & \deepred{$52.94 _{\pm 3.27}$} & \deepred{$50.79 _{\pm 2.63}$} & \deepred{$50.70 _{\pm 2.61}$} & \deepred{$50.60 _{\pm 2.58}$} & \deepreds{$49.51 _{\pm 4.57}$} & \deepreds{$50.84 _{\pm 3.90}$} & \deepreds{$0.97$M} & \deepreds{$3.76$G} \\
& Improve. & \deepred{$ +2.42 $} & \deepred{$ + 2.60 $} & \deepred{$ +2.48 $} & \deepred{$ +2.44 $} & \deepreds{$ +1.23 $} & \deepreds{$ + 0.13 $} & \deepred{$8\times $} & \deepred{$23\times $} \\
\hline 
\multirow{11}{*}{\begin{tabular}{l}
PTB \\
\end{tabular}} 
& Mamba & $81.00 _{\pm 1.41}$ & $84.48 _{\pm 2.37}$ & $72.62 _{\pm 1.67}$ & $74.86 _{\pm 1.91}$ & $91.16 _{\pm 1.86}$ & $90.40_{\pm 2.05}$ & $0.76$M & $1.54$G \\
& S-Mamba & $82.60 _{\pm 1.32}$ & $85.39 _{\pm 1.77}$ & $75.28 _{\pm 2.00}$ & $77.61 _{\pm 2.03}$ & $92.19 _{\pm 0.10}$ & $91.62 _{\pm 1.13}$ & $1.07$M & $2.18$G \\
& Transformer & $77.10 _{\pm 2.27}$ & $79.80 _{\pm 2.16}$ & $ 67.31_{\pm 3.61}$ & $68.57 _{\pm 4.38}$ & $90.02 _{\pm 2.57}$ & $86.15 _{\pm 2.37}$ & $0.88$M & $48.44$G \\
& Nonformer & $78.76 _{\pm 1.80}$ & $82.60_{\pm 1.91}$ & $69.35_{\pm 2.66}$ & $71.11 _{\pm 3.11}$ & $89.98_{\pm 1.25}$ & $86.78_{\pm 2.02}$ & $0.96$ M & $48.46$ G \\
& Crossformer & $84.35_{\pm 2.59}$ & $87.04_{\pm 1.02}$ & $ 77.81_{\pm 4.38}$ & $80.05_{\pm 4.22}$ & $91.98_{\pm 1.54}$ & $91.62 _{\pm 1.45}$ & $5.24$ M & $29.21 $ G \\
& PatchTST & $77.56_{\pm 1.46}$ & $80.30_{\pm 1.13}$ & $ 68.00_{\pm 2.33}$ & $69.48_{\pm 2.75}$ & $89.54_{\pm 2.24}$ & $84.48 _{\pm 3.20}$ & $0.94$ M & $60.66$ G \\
& iTransformer & $82.88 _{\pm 2.38}$ & $87.07 _{\pm 2.64}$ & $ 75.02_{\pm 3.26}$ & $77.52 _{\pm 3.59}$ & $90.97 _{\pm 1.40}$ & $90.63_{\pm 1.68}$ & $0.84$ M & $1.64$ G \\
& Medformer & $77.89 _{\pm 2.53}$ & $81.38 _{\pm 1.64}$ & $68.23 _{\pm4.16 }$ & $69.62 _{\pm 4.82}$ & $93.06 _{\pm 0.59}$ & $90.74 _{\pm 0.86}$ & $6.10$M & $49.77$G \\
\cmidrule(lr){2-10} 
& BioMamba (Ours) & \deepred{$84.53 _{\pm 3.12}$} & \deepred{$87.50 _{\pm 2.20}$} & \deepred{$77.86 _{\pm 4.88}$} & \deepred{$80.18 _{\pm 4.85}$} & \deepred{$95.14 _{\pm 0.61}$} & \deepred{$94.30 _{\pm 1.10}$} & \deepreds{$0.82$M} & \deepreds{$4.04$G}  \\
& Improve. & \deepred{$ +6.64 $} & \deepred{$ +6.12 $} & \deepred{$ +9.63 $} & \deepred{$ +10.56 $} & \deepred{$ +2.08 $} & \deepred{$ +3.56 $} & \deepred{$ 7\times $} & \deepred{$ 12\times $}\\
\bottomrule
\end{tabular}
}
\caption{BioMamba achieves state-of-the-art biosignals classification performance in the five datasets, evaluated across six distinct metrics, all with fewer than 1 M parameters, outpacing previous models by a significant margin. It also reduces the computational cost ( FLOPs ) from 2x to 23x compared to Medformer~\cite{wang2024medformer}. The best results are in \deepred{bold}.}
\label{tab: overall_performance}
\end{table*}

\subsection{Overall Comparison}

In Table~\ref{tab: overall_performance}, we present the performance and effectiveness of BioMamba alongside eight benchmark methods in the biosignal classification task. BioMamba demonstrates superior performance across all six evaluation metrics on five out of six datasets, achieving the highest scores in Accuracy, Precision, Recall, F1 Score, AUROC, and AUPRC. Compared to Mamba-based methods, BioMamba reaches this level with a comparable parameter count. Against Attention-based methods, BioMamba consistently outperforms across all datasets across five evaluation metrics. For example, it achieves an accuracy of 96.77\%, surpassing Medformer by 8.46\% on the TDBrain dataset. We can also see from Figure~\ref{fig: radar_average_results} that BioMamba persistently outperforms previous methods with an average of six datasets results. 
Additionally, Medformer~\cite{wang2024medformer} and Crossformer~\cite{zhang2023crossformer} perform well across the six datasets, benefiting from their cross-channel learning strategy.

In terms of computational efficiency, BioMamba shows notable capability, with a smaller parameter count than the Attention-based methods. For instance, BioMamba requires only 0.73M parameters compared to Medformer’s~\cite{wang2024medformer} 7.35M on the STEW dataset, achieving a $10 \times$ reduction in computational cost.
The high efficiency in computational resources enables BioMamba to capture long temporal dependencies within a limited computation budget.
An overview of average performance across all six metrics is provided in Table~\ref{tab: average_rank}.

\begin{table*}[htpb]
\centering
\resizebox{0.86\textwidth}{!}{%
\begin{tabular}{c|cc|cc|cc|cc}
\toprule 
\multirow{2}{*}{Datasets}  & \multicolumn{2}{c|}{ADFTD} & \multicolumn{2}{c|}{PTB-XL} & \multicolumn{2}{c|}{UCI-HAR}  & \multicolumn{2}{c}{FLAAP} \\
& \multicolumn{2}{c|}{(3-Classes)} & \multicolumn{2}{c|}{(5-Classes)} & \multicolumn{2}{c|}{(6-Classes)}  & \multicolumn{2}{c}{(10-Classes)} \\
\hline
\diagbox[innerleftsep=0.1cm, innerrightsep=0.2cm]{Models}{Performance}   & Accuracy & F1 score & Accuracy & F1 score & Accuracy & F1 score & Accuracy & F1 score  \\
\hline
Mamba  & $50.24 _{\pm 1.18}$  & $45.69 _{\pm 0.56}$  & $69.36 _{\pm 0.32}$  & $56.08 _{\pm 0.42}$  & $89.51 _{\pm 0.32}$ & $89.22 _{\pm 0.33}$ & $67.45 _{\pm 0.36}$  & $66.49 _{\pm 0.38}$ \\
S-Mamba  & $50.52 _{\pm 0.60}$   & $46.12 _{\pm 0.32}$   & $69.55 _{\pm 0.25}$  & $56.36 _{\pm 0.23}$  & $90.61 _{\pm 0.12}$  & $90.37 _{\pm 0.11}$ & $69.18 _{\pm 0.71}$ & $68.14 _{\pm 0.69}$ \\
TCN & $50.90 _{\pm 1.62}$  & $47.46 _{\pm 1.66}$  & \deepred{$73.42 _{\pm 0.79}$} & \deepred{$62.63 _{\pm 0.39}$} & $93.13 _{\pm 1.32}$  & $93.13 _{\pm 1.31}$  & $67.87 _{\pm 4.17}$  & $66.66 _{\pm 4.23}$ \\
Transformer  & $50.86 _{\pm 1.42}$   &  \deepred{$48.09 _{\pm 1.34}$}  & $70.43 _{\pm 0.45}$  & $58.66 _{\pm 0.45}$  & $89.94 _{\pm 2.12}$ & $89.83 _{\pm 2.16}$   & $76.07 _{\pm 0.67}$  & $75.62 _{\pm 0.63}$ \\
Crossformer & $49.93 _{\pm 1.52}$  & $45.32 _{\pm 0.96}$  & $73.38 _{\pm 0.62}$ & $62.60 _{\pm 0.89}$ & $90.36 _{\pm 0.76}$  & $90.41 _{\pm 0.76}$ & $76.34 _{\pm 0.43}$ & $76.10 _{\pm 0.44}$ \\
PatchTST & $41.91 _{\pm 1.19}$  & $40.61 _{\pm 2.20}$  & $73.20 _{\pm 0.20}$ & $62.40 _{\pm 0.56}$ & $87.19 _{\pm 0.49}$ & $87.55 _{\pm 0.60}$  & $56.21 _{\pm 0.69}$  & $55.24 _{\pm 0.88}$ \\

Medformer  &  \deepred{$51.54 _{\pm 1.09}$}   & $46.42 _{\pm 1.52}$ & $72.75 _{\pm 0.09}$  & $61.42 _{\pm 0.20}$  & $91.80 _{\pm 0.62}$  & $91.78 _{\pm 0.65}$ & $77.50 _{\pm 0.67}$  & $77.32 _{\pm 0.78}$ \\
\hline 
BioMamba (Ours) &  $48.08 _{\pm 0.30}$  & $43.93 _{\pm 0.39}$  & $71.08 _{\pm 0.12}$  & $58.40 _{\pm 0.41}$ & \deepred{$94.42 _{\pm 0.09}$} &  \deepred{$94.42 _{\pm 0.09}$}  & \deepred{$78.51 _{\pm 0.36}$} & \deepred{$78.32 _{\pm 0.39}$} \\
\bottomrule
\end{tabular}
}
\caption{Further biosignals classification results. The best scores are in \deepred{bold}. BioMamba outperforms baselines in the two human activity recognition tasks.}
\label{tab: performance_1}
\end{table*}

\subsection{Further Study}

As shown in Table~\ref{tab: performance_1}, we further study BioMamba in multiclass classification tasks, including brain disease detection, heart disease classification, and human activity recognition. Our BioMamba outperforms the baselines in two human activity recognition tasks. Additionally, Medformer~\cite{wang2024medformer} and Transformer~\cite{vaswani2017attention} demonstrate strong performance in the ADFTD~\cite{miltiadous2023dataset, miltiadous2023dice} task, while TCN~\cite{bai2018empirical} outperforms the others in the PTB-XL~\cite{wagner2020ptb}  task,
details in Appendix~\ref{sec: ad_experiments}.

\subsection{Efficient Training} 
We evaluate the training efficiency of BioMamba against eight baselines across six diverse tasks, presenting both training time per epoch and GPU memory consumption (see Table~\ref{tab: model_efficiency}).
In terms of training time per epoch, BioMamba achieves acceptable training times relative to the baselines while maintaining top-1 accuracy across all datasets. While Medformer~\cite{wang2024medformer} presents a long training time due to its multi-granularity patching approach. Regarding GPU memory consumption, BioMamba achieves $1\times$-$10\times$ improvement of Medformer across the six different tasks. Notably, 
with the variable-wise embedding, iTransformer~\cite{liu2023itransformer} demonstrates effective learning across all Attention-based methods in GPU memory consumption. In addition, the Mamba-based baselines, including vanilla Mamba~\cite{gu2023mamba} demonstrate a compelling advantage in training times. This suggests that our BioMamba offers a more efficient approach for biosignal processing than Medformer~\cite{wang2024medformer}, boosting its real-world applicability.

\subsection{Ablation Studies}
We perform comprehensive ablation studies on the key components and hyperparameter choices of BioMamba, reporting performance across six datasets in Appendix~\ref{sec: Ablation Studies}.


\subsection{Analysis}
\label{sec: analysis}

We analyze BioMamba from three aspects: reliability, efficiency, and generality, These three aspects correspond to the key contributions of BioMamba.
\textbf{(1) Reliability:}
As shown in Table~\ref{tab: average_rank}, our BioMamba consistently achieves high performance, validated through six classification evaluation metrics that underscore its robustness and reliability across diverse tasks. With an average improvement of 5\%–7\% over Medformer~\cite{wang2024medformer}, BioMamba demonstrates significant advancements in classification capability, establishing a new state-of-the-art benchmark in biosignal analysis.
\textbf{(2) Efficiency:} Table~\ref{tab: characteristics} highlights the \textbf{Model Efficiency} and \textbf{Training Efficiency} characteristics of BioMamba in comparison with eight baseline models. This analysis reveals that BioMamba, alongside Vanilla Mamba~\cite{gu2023mamba} and iTransformer~\cite{liu2023itransformer}, achieves computational efficiency, effectively reducing the model size and GPU resource usage compared to alternative approaches. We also analyze the details of computational complexity, see Appendix~\ref{sec: complexity}.
\textbf{(3) Generality:} We evaluate BioMamba on ten clinical tasks (see Table~\ref{tab: all_datasets} and Table~\ref{tab: performance_1}), emphasizing its capability for precise classification in diverse settings with efficient learning.  This advancement not only strengthens the Mamba model family for biosignal analysis but also enhances its practical applicability. BioMamba demonstrates adaptability and effectiveness across a wide range of domains and applications.

\begin{table}[htpb]
\centering
\resizebox{\columnwidth}{!}{
\begin{tabular}{lccccc}
\toprule
\multirow{2}{*}{Models}   & \multicolumn{2}{c} {Training Efficiency}   & \multicolumn{2}{c} {Model Efficiency} & \multirow{2}{*}{\parbox{2cm}{Classification \\ Performance}}\\
 \cmidrule(lr){2-3} \cmidrule(lr){4-5}    
 & Training Time & GPU Memory & Model Size & Operations    \\
\hline 
Mamba             & \checkmark & \checkmark   & \checkmark  & \checkmark\\
S-Mamba           & \checkmark & \checkmark  &             & \checkmark\\
Transformer       & \checkmark &              & \checkmark  &            \\
Nonformer         & \checkmark &              & \checkmark  & \\
Crossformer       & \checkmark &            &             & \\
PatchTST          & \checkmark &              & \checkmark  &  \\
iTransformer      & \checkmark & \checkmark    & \checkmark  & \checkmark\\
Medformer         &          &              &             & \\
 \rowcolor{purple!20}
BioMamba (Ours)   & \checkmark & \checkmark & \checkmark  & \checkmark & \checkmark  \\
\bottomrule
\end{tabular}
}
\caption{Conclusion of efficiency and performance between Existing Methods and BioMamba.}
\label{tab: characteristics}
\end{table}


\section{Conclusion}
\label{sec: conclusion}

This paper addresses the limitations of existing Attention-based and Mamba-based models in biosignal classification tasks, specifically targeting issues of inefficient learning, high computational overhead, and suboptimal performance. We propose a novel method, BioMamba, which leverages the Spectro-Temporal Embedding in Bidirectional Mamba with the Sparse Feed Forward policy. 
Our extensive experiments demonstrate that BioMamba achieves new state-of-the-art performance with high learning efficiency on most biosignal classification benchmarks.
Our BioMamba enhances the Mamba family in biosignal analysis, promoting better utilization of real-world scenarios, and making it practical for wearable and portable medical equipment.

\bibliography{icml2024}
\bibliographystyle{icml2024}
\clearpage

\appendix
\onecolumn

\clearpage
\begin{center}
    \large \bf {Appendix of BioMamba}
\end{center}

\section{Datasets of Experimental Setups}

\begin{figure*}[b!]
    \centering
    \includegraphics[width=1\linewidth]{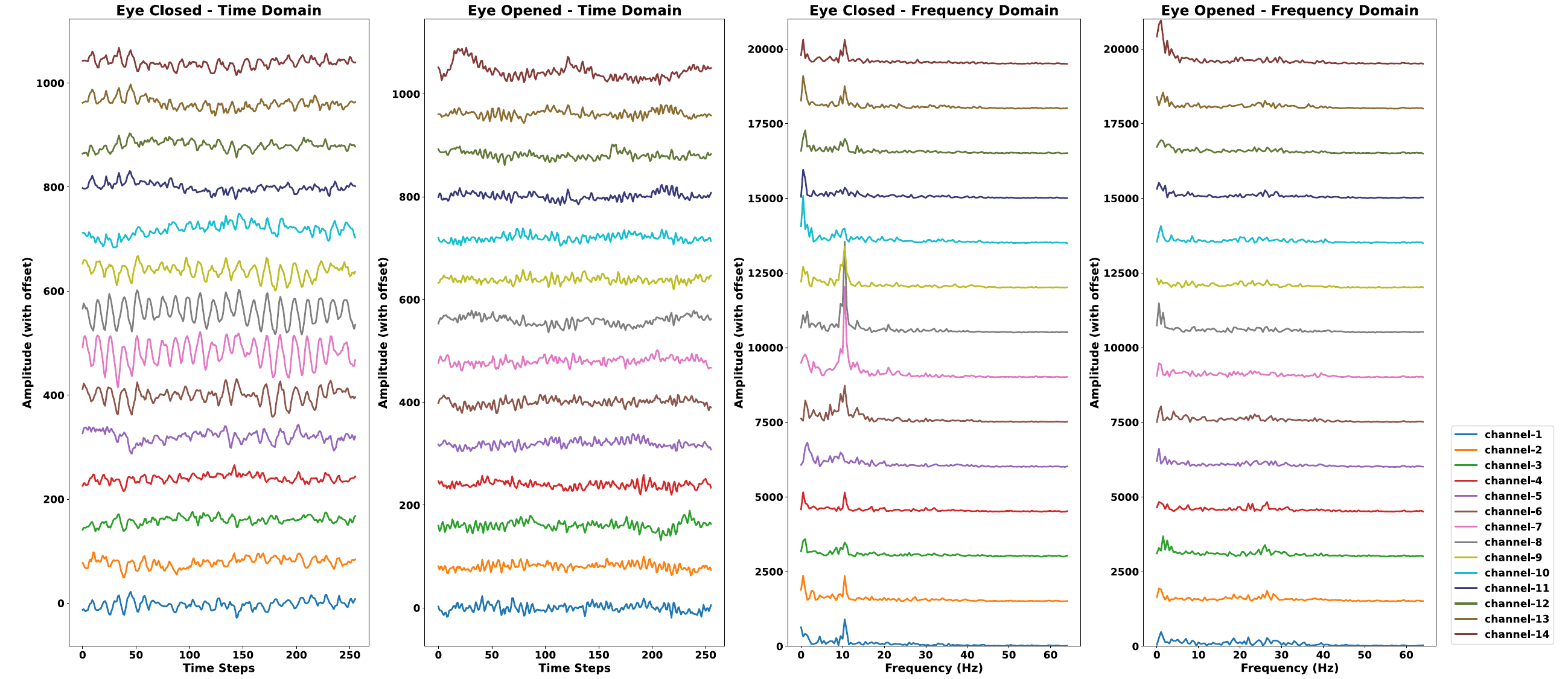} 
    \caption{Visualization of the CrowdSource dataset in both the time domain and frequency domain. To enhance the clarity of the channel information, we apply normalization and offset adjustments to the original data.}
    \label{fig: crowdsource_dataset}
\end{figure*}

\subsection{Data Description}
\label{sec: datasets}
\textbf{APAVA} dataset~\cite{escudero2006analysis} is an EEG dataset with binary-labeled samples indicating the presence of Alzheimer’s disease. It comprises two classes across 23 subjects, including 12 Alzheimer’s disease patients and 11 healthy controls. On average, each subject has $30.0 \pm$ 12.5 trials, with each trial being a 5-second time sequence consisting of 1280 timestamps across 16 channels. 
Following the Medformer, we employ the subject-wise setup, samples with subject IDs $\{15,16,19,20\}$ and $\{1,2,17,18\}$ are allocated to the validation and test sets, respectively. The remaining samples are organized into the training set.

\textbf{TDBrain}~\cite{van2022two} is an EEG dataset in which each sample is assigned a binary label indicating whether the subject has Parkinson’s disease. The dataset comprises brain activity recordings from 1274 subjects across 33 channels, with each subject undergoing eyes open/closed trials. A total of 60 labels are provided, with each subject potentially having multiple labels to denote multiple co-existing conditions. As same as Medformer, we utilize a subset of the dataset containing 25 subjects with Parkinson's disease and 25 healthy controls, all under the eyes-closed condition. A subject-wise setup is used for training, validation, and test splits: samples from subjects with IDs $\{18,19,20,21,46,47,48,49\}$ are assigned to the validation set, and those from subjects with IDs $\{22,23,24,25,50,51,52,53\}$ are placed to the test set. The remaining samples are reserved for training.

\textbf{Crowdsourced} dataset~\cite{williams2023crowdsourced} was collected while participants engaged in a resting state task, alternating between two-minute periods with eyes open and eyes closed. Among 60 participants, 13 successfully completed both conditions using 14-channel EPOC+, EPOC X, and EPOC devices. The data was originally recorded at 2048 Hz and subsequently downsampled to 128 Hz. Raw EEG data for these 13 participants, along with preprocessing, analysis, and visualization scripts, are publicly accessible on the Open Science Framework (OSF).

\textbf{STEW} dataset~\cite{lim2018stew} comprises raw EEG recordings from 48 participants using a 14-channel Emotiv EPOC headset during a multitasking workload experiment with the SIMKAP multitasking test. Baseline brain activity was also recorded while subjects were at rest before the test. Data was captured at a sampling rate of 128 Hz across 14 channels, yielding 2.5 minutes of EEG recordings per participant.

\textbf{DREAMER}~\cite{katsigiannis2017dreamer} is a multimodal database containing electroencephalogram (EEG) and electrocardiogram (ECG) signals recorded during affect elicitation through audio-visual stimuli, using a 14-channel Emotiv EPOC headset. In this study, we apply the task on the EEG data.

\textbf{PTB} dataset~\cite{goldberger2000physiobank} is a public ECG time-series dataset containing recordings from 290 subjects across 15 channels, with a total of 8 labels indicating 7 types of heart disease and 1 healthy control. The original sampling rate is 1000 Hz. 
For a fair comparison, we use the same preprocessing as Medformer. The ECG signals are downsampled to 250 Hz and normalized using standard scalers. Then, we identify R-Peak intervals across all channels, removing outliers, and sampling each heartbeat from its R-Peak position. For training, validation, and test splits, we also employ a subject-wise setup, assigning 60\%, 20\%, and 20\% of subjects and their corresponding samples to the training, validation, and test sets, correspondingly.

\subsection{A Biosignal Example}
We present the characteristics of six datasets in Table~\ref{tab: all_datasets} engaging with six different clinical tasks, including Alzheimer's Disease Classification, Parkinson's Disease Detection, Eyes Open/Close Detection, Mental Workload Classification, Emotion Detection, and Myocardial Infarction Detection. In Figure~\ref{fig: crowdsource_dataset}, we display the original Crowdsourced dataset information in the temporal domain and frequency domain, containing 14-channel EEG data, with each segment preprocessed to 256 time steps.
We can directly observe the frequency differences between closed and open eyes in the frequency domain, which confirms the effectiveness of our Spectro-Temporal Embedding strategy.

\section{BioMamba Block Framework}
\label{sec:BioMamba_Block_Framework}

\subsection{Network Architecture}

\begin{figure*}[htpb]
    \centering
    \includegraphics[width=0.8\linewidth]{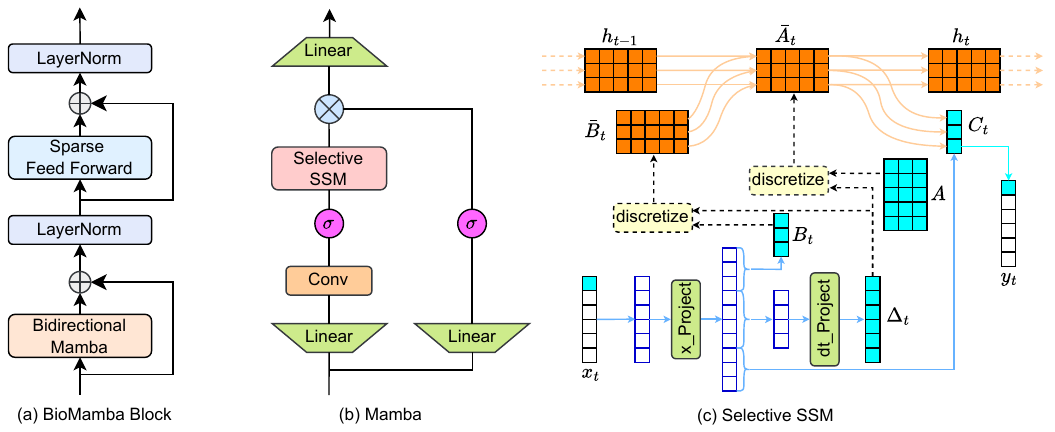} 
    \caption{The detailed structure of the BioMamba block, Mamba process, and the Selective SSM mechanism. }
    \label{fig: method_blocks}
\end{figure*}

In~\ref{sec:bidirectional_mamba} we introduce the architecture of our Bidirectional Mamba layer, which consists of two Mamba processes. Figure~\ref{fig: method_blocks} is a detailed illustration of the architectures of the Bidirectional Mamba, incorporating the pipeline of selective SSM mechanism. Specifically, Figure~\ref{fig: method_blocks} (c) illustrates the pipeline of the Selective SSM. As we can see,
the selection mechanism allows the input to participate in updating the learning parameters $(\Delta_t, \boldsymbol{B}, \boldsymbol{C})$, enabling the model to adapt with the information and granting it the ability to select relevant features. This mechanism efficiently extracts essential information from input sequence elements and captures long-range dependencies that scale with sequence length while maintaining linear computational complexity (see Table~\ref{tab: complexity}) for handling extended sequences. We present the pseudo-code for the Bidirectional Mamba framework in Algorithm~\ref{alg: mamba_and_ssm}, providing a detailed illustration of the Bidirectional Mamba process with the Selective SSM Mechanism.

\begin{algorithm}[htpb]
   \caption{The Bidirectional Mamba Process with Selective SSM Mechanism}
   \label{alg: mamba_and_ssm}
\begin{algorithmic}
\small
   \STATE \textbf{Input:} $\mathbf{Z}^0 = \left[\mathbf{z}^1, \mathbf{z}^2, \ldots, \mathbf{z}^B\right]:(B, E, D)$
   \STATE \textbf{Output:} $\mathbf{Z}^m = \left[{\mathbf{z}}^1, {\mathbf{z}}^2, \ldots, {\mathbf{z}}^B\right]:(B,E,D)$
   \FOR{$m$ \textbf{in layers}}
    \STATE $\operatorname{Bidrirectional~Mamba~:}$ 
    \STATE  \quad $\operatorname{Mamba(+)~:}$ 
 \STATE \quad \quad $  \mathbf{Z^{m-1}_{b 1}}=\operatorname{Selective-SSM}(\operatorname{\sigma}(\operatorname{Conv}(\operatorname{Linear}(   \mathbf{Z^{m-1}} )))) \quad  /* \sigma~represents~SiLU~activation~function. */$
     \STATE \quad \quad  $  \mathbf{Z^{m-1}_{b 2}}=\operatorname{\sigma}(\operatorname{Linear}( \mathbf{Z^{m-1}} ))   $
     \STATE \quad \quad  $ \mathbf{Z^{m-1}_1}=\operatorname{Linear}\left( \mathbf{Z^{m-1}_{b 1}} \odot  \mathbf{Z^{m-1}_{b 2}}\right)  \quad  /* \odot ~represents~element-wise~multiplication. */  $
     \STATE  \quad $\operatorname{Mamba(-)~:}$ 
     \STATE \quad \quad $  \mathbf{Z^{m-1}_{b 1}}=\operatorname{Selective-SSM}(\operatorname{\sigma}(\operatorname{Conv}(\operatorname{Linear}(\operatorname{Reverse}(  \mathbf{Z^{m-1}} ))))) $
     \STATE \quad \quad  $  \mathbf{Z^{m-1}_{b 2}}=\operatorname{\sigma}(\operatorname{Linear}(\operatorname{Reverse}(  \mathbf{Z^{m-1}} ))) $
     \STATE \quad \quad  $\mathbf{Z^{m-1}_2} =\operatorname{Reverse}(\operatorname{Linear}\left( \mathbf{Z^{m-1}_{b 1}} \odot  \mathbf{Z^{m-1}_{b 2}}\right)) $

   \STATE $\mathbf{Z^{m-1}}:(B, E, D) \leftarrow \operatorname{LN}((\mathbf{Z^{m-1}_1} + \mathbf{Z^{m-1}_2}) + \mathbf{Z^{m-1}})$
   \STATE $\mathbf{Z^{m-1}_s}:(B, E, D) \leftarrow \operatorname{Sparse~Feed~Forward}(\mathbf{Z^{m-1}})$
  \STATE $\mathbf{Z^{m}}:(B, E, D) \leftarrow \operatorname{LN}(\mathbf{Z^{m-1}_s}+ \mathbf{Z^{m-1}})$
   \ENDFOR 
\end{algorithmic}
\end{algorithm}

\subsection{Baselines}
\label{sec: Baselines}
\textbf{Mamba}~\cite{gu2023mamba} has demonstrated excellent performance in sequence modeling by introducing a data-dependent selection mechanism based on S4, which efficiently filters specific inputs and captures long-range context that scales with sequence length. The raw code is available at~\textbf{\url{https://github.com/state-spaces/mamba}}.

\textbf{S-Mamba}~\cite{wang2024mamba} utilizes Bidirectional Mamba to set new benchmarks in time-series forecasting, achieving state-of-the-art performance with considerably reduced computational cost relative to Attention-based approaches. The code is available at~\textbf{\url{{https://github.com/wzhwzhwzh0921/S-D-Mamba}}}.

\textbf{Vanilla Transformer}~\cite{vaswani2017attention} is presented in "Attention is All You Need." It can also be utilized in time-series by encoding each timestamp of all channels as an attention token. The PyTorch version of the code can be accessed at~\textbf{\url{https://github.com/jadore801120/attention-is-all-you-need-pytorch}}.

\textbf{Nonformer}~\cite{liu2022non} tackles the challenges of non-stationarity in time-series forecasting, uncovering its substantial impact on performance. It presents a de-stationary attention module and utilizes normalization and denormalization before and after training to alleviate over-rationalization. The code can be accessed at~\textbf{\url{https://github.com/thuml/Nonstationary_Transformers}}.

\textbf{Crossformer}~\cite{zhang2023crossformer} presents a single-channel patching method for token embedding, utilizing a two-stage self-attention mechanism to grasp temporal features and channel correlations effectively. A router mechanism further enhances time and space efficiency in the cross-dimension stage. The code can be accessed at~\textbf{\url{https://github.com/Thinklab-SJTU/Crossformer}}.

\textbf{PatchTST}~\cite{nie2022time} improves time-series forecasting by dividing sequences into patches, expanding input length while reducing redundancy. This method extends the receptive field, significantly enhancing forecasting performance. The code can be accessed at~\textbf{\url{https://github.com/yuqinie98/PatchTST}}.

\textbf{iTransformer}~\cite{liu2023itransformer} questions the traditional token embedding approach in time-series forecasting by encoding the entire series of channels into a single token. This method also inverts the dimensions in other transformer modules, including layer normalization and feed-forward networks. The code can be accessed at~\textbf{\url{https://github.com/thuml/iTransformer}}.

\textbf{Medformer}~\cite{wang2024medformer} introduces a multi-granularity patching transformer and two-stage multi-granularity self-attention for learning features and correlations, achieving promising results for medical time-series classification. The raw code can be accessed at~\textbf{\url{https://github.com/DL4mHealth/Medformer}}.

\begin{table*}[htpb]
\centering
\resizebox{0.8\textwidth}{!}{%
\begin{tabular}{c|cccccc}
\toprule
Hyperparameters & APAVA & TDBrain & Crowdsourced &  STEW& DREAMER & PTB  \\
\hline 
Frequency Resolution &  [200,~~50] &  [256,~~50] &  [128,~100] &  [256,~~50]  &[256,~~50] &  [256,~~50] \\
Sparsity             & 0.3 & 0.7 &  0.7& 0.9 &0.3 & 0.7  \\
BioMamba Blocks      & 6 & 6 & 6 & 6  & 6&  6 \\
Hidden Dimension     & 128 & 128 & 128  & 128 & 128 & 128  \\
Batch Size           & 32 & 32 & 32 & 64          & 128 &  128 \\
Learning Rate  & 5e-5 & 5e-5 &5e-5  &   5e-5       &5e-5 & 5e-5  \\
Training Epochs & 100 &  100&  100&100& 100& 100  \\
\bottomrule
\end{tabular}
}
\caption{Hyperparameters for BioMamba.}
\label{tab: hyperparameters}
\end{table*}

\section{Implementation Details}
\label{sec: implementation_details}

\begin{table*}[htpb]
\centering
\resizebox{0.9\textwidth}{!}{%
\begin{tabular}{cccccccc|cc}
\toprule
Datasets & Embedding & Accuracy & Precision & Recall & F1 score & AUROC & AUPRC & Params (M) & FLOPs (G)  \\
\hline 
\multirow{3}{*}{\begin{tabular}{l}
APAVA \\
\end{tabular}} 
& w/o PSE  & $74.84 _{\pm 1.73}$ & $75.92 _{\pm 1.64}$ & $71.51 _{\pm 2.13}$ & $72.07 _{\pm 2.22}$ & $85.90 _{\pm 1.71}$ & $84.76 _{\pm 1.84}$ & $0.95$M & $0.54$G \\
& w/o TDE  & $83.59 _{\pm 2.29}$ & $83.84 _{\pm 2.31}$ & $82.03 _{\pm 2.60}$ & $82.61 _{\pm 2.53}$ & $92.62 _{\pm 1.41}$ & $92.17 _{\pm 1.65}$ & $0.94$M & $1.06$G \\
& BioMamba & \deepreds{$84.95 _{\pm 1.35}$} & \deepreds{$85.72 _{\pm 1.95}$} & \deepreds{$83.15 _{\pm 1.13}$} & \deepreds{$83.95 _{\pm 1.32}$} & \deepreds{$93.79 _{\pm 1.39}$} & \deepreds{$93.52 _{\pm 1.40}$} & \deepreds{$0.97$M} & \deepreds{$1.61$G} \\
\hline
\multirow{3}{*}{\begin{tabular}{l}
TDBrain \\
\end{tabular}} 
& w/o PSE  & $73.19 _{\pm 1.59}$ & $73.45 _{\pm 1.64}$ & $73.19 _{\pm 1.59}$ & $73.11 _{\pm 1.59}$ & $81.72 _{\pm 1.28}$ & $81.70 _{\pm 1.28}$ & $0.80$M & $1.12$G \\
& w/o TDE  & $94.40 _{\pm 3.55}$ & $94.46 _{\pm 3.53}$ & $94.40 _{\pm 3.55}$ & $94.39 _{\pm 3.55}$ & $98.48 _{\pm 1.42}$ & $98.51 _{\pm 1.37}$ & $0.78$M & $1.10$G \\
& BioMamba & \deepreds{$96.77 _{\pm 1.94}$} & \deepreds{$96.90 _{\pm 1.71}$} & \deepreds{$96.77 _{\pm 1.94}$} & \deepreds{$96.77 _{\pm 1.95}$} & \deepreds{$99.44 _{\pm 0.49}$} & \deepreds{$99.42 _{\pm 0.51}$} & \deepreds{$0.83$M} & \deepreds{$2.22$G} \\
\hline
\multirow{3}{*}{\begin{tabular}{l}
Crowdsourced \\
\end{tabular}} 
& w/o PSE  & $77.00 _{\pm 1.06}$ & $77.95 _{\pm 1.13}$ & $76.99 _{\pm 1.06}$ & $76.80 _{\pm 1.08}$ & $87.26 _{\pm 1.66}$ & $86.99 _{\pm 1.84}$ & $0.71$M & $0.47$G \\
& w/o TDE  & $89.64 _{\pm 1.32}$ & $89.83 _{\pm 1.21}$ & {$89.63 _{\pm 1.32}$} & $89.62 _{\pm 1.33}$ & $96.63 _{\pm 0.49}$ & $96.67 _{\pm 0.52}$ & $0.69$M & $0.93$G \\
& BioMamba & \deepreds{$89.84 _{\pm 0.72}$} & \deepreds{$90.04 _{\pm 0.75}$} & \deepreds{$89.83 _{\pm 0.72}$} & \deepreds{$89.82 _{\pm 0.71}$} & \deepreds{$96.88 _{\pm 0.34}$} & \deepreds{$96.97 _{\pm 0.33}$} & \deepreds{$0.81$M} & \deepreds{$1.40$G} \\
\hline
\multirow{3}{*}{\begin{tabular}{l}
STEW \\
\end{tabular}} 
& w/o PSE  & $67.72 _{\pm 0.79}$ & $68.67 _{\pm 1.03}$ & $67.72 _{\pm 0.79}$ & $67.30 _{\pm 0.73}$& $75.61 _{\pm 1.30}$ & $74.58 _{\pm 1.21}$ & $0.71$M & $0.95$G \\
& w/o TDE  & \deepreds{$79.76 _{\pm 0.56}$} & \deepreds{$79.78 _{\pm 0.55}$} & \deepreds{$79.76 _{\pm 0.56}$} & \deepreds{$79.75 _{\pm 0.56}$} & $87.43 _{\pm 0.52}$ & $87.18 _{\pm 0.51}$ & $0.70$M & $0.93$G \\
& BioMamba & {$79.60 _{\pm 1.00}$} & {$79.65 _{\pm 1.03}$} & {$79.60 _{\pm 1.00}$} & {$79.59 _{\pm 0.99}$} & \deepreds{$87.44 _{\pm 0.56}$} & \deepreds{$87.27 _{\pm 0.53}$} & \deepreds{$0.73 $M} & \deepreds{$1.88$G} \\
\hline
\multirow{3}{*}{\begin{tabular}{l}
DREAMER \\
\end{tabular}} 
& w/o PSE  & $51.11 _{\pm 3.21}$ & $48.12 _{\pm 3.67}$ & $48.24 _{\pm 3.44}$ & $48.04 _{\pm 3.58}$ & $47.97 _{\pm 4.03}$ & $50.66 _{\pm 2.38}$ & $0.95$M & $1.90$G \\
& w/o TDE  & $52.89 _{\pm 4.92}$ & $50.59 _{\pm 5.07}$ & $50.57 _{\pm 5.11}$ & $50.41 _{\pm 5.03}$ & $48.05 _{\pm 4.09}$ & $47.63 _{\pm 2.39}$ & $0.93$M & $1.87$G \\
& BioMamba & \deepreds{$52.94 _{\pm 3.27}$} & \deepreds{$50.79 _{\pm 2.63}$} & \deepreds{$50.70 _{\pm 2.61}$} & \deepreds{$50.60 _{\pm 2.58}$} & \deepreds{$49.51 _{\pm 4.57}$} & \deepreds{$50.84 _{\pm 3.90}$} & \deepreds{$0.97$M} & \deepreds{$3.76$G} \\
\hline
\multirow{3}{*}{\begin{tabular}{l}
PTB \\
\end{tabular}} 
& w/o PSE  & $80.30 _{\pm 1.92}$ & $84.05 _{\pm 1.13}$ & $71.60 _{\pm 3.05}$ & $73.65 _{\pm 3.41}$ & $93.23 _{\pm 0.58}$ & $91.11 _{\pm 0.75}$ & $0.80$M & $2.04$G \\
& w/o TDE  & $84.10 _{\pm 1.43}$ & \deepreds{$87.70 _{\pm 1.48}$} & $76.99 _{\pm 2.19}$ & $79.53 _{\pm 2.23}$ & $91.80 _{\pm 2.76}$ & $91.57 _{\pm 2.60}$ & $0.78$M & $2.00$G \\
& BioMamba & \deepreds{$84.53 _{\pm 3.12}$} & {$87.50 _{\pm 2.20}$} & \deepreds{$77.86 _{\pm 4.88}$} & \deepreds{$80.18 _{\pm 4.85}$} & \deepreds{$95.14 _{\pm 0.61}$} & \deepreds{$94.30 _{\pm 1.10}$} & \deepreds{$0.82$M} & \deepreds{$4.04$G}  \\
\bottomrule
\end{tabular}
}
\caption{Ablation study on different embedding configurations to analyze the impact of Patched Spectral Embedding (PSE) and Temporal Domain Embedding (TDE). 
Configurations \textbf{without (w/o)} PSE or TDE are compared to the default model.  We can find that the PSE significantly boosts the classification performance of BioMamba. The best results are in \textbf{bold}.}
\label{tab: ablation_of_embedding}
\end{table*}

\subsection{Evaluation Metrics}

\textbf{Accuracy} is a core metric for evaluating classification models, representing the proportion of correctly predicted samples out of the total samples. It is applicable across both binary and multi-class classification tasks.

\textbf{Precision} measures the proportion of correctly predicted positive instances among all instances predicted as positive, indicating the model’s accuracy in identifying true positives.

\textbf{Recall} represents the proportion of correctly identified positive instances out of all actual positive instances, measuring the model's effectiveness in capturing true positives comprehensively.

\textbf{F1 Score} is the harmonic mean of precision and recall, making it especially valuable when a balance between these metrics is essential. In this paper, the weighted F1 score is employed for both binary and multi-class classification, representing a weighted average of each class's individual F1 score, with weights proportional to the number of samples per class.

\textbf{AUROC} (Area Under the Receiver Operating Characteristic Curve) condenses the ROC curve into a single value, representing model performance across multiple thresholds in binary classification. A higher AUROC indicates a stronger ability of the model to distinguish between the two classes.

\textbf{AUCPR} (Area Under the Precision-Recall Curve) represents the area under the precision-recall curve for binary classification, offering a more insightful performance measure for imbalanced data compared to AUROC. It highlights the model's effectiveness in maintaining high precision and recall across varying thresholds.

\subsection{Implementation  Setups}
\label{subsec: setups}
We implement BioMamba along with all baseline methods using PyTorch~\cite{paszke2019pytorch}. All methods are optimized using the Adam optimizer~\cite{kingma2014adam}. For Mamba-based models, we set the learning rate to $\{5e-5\}$, while for Attention-based methods, it is set to $\{1e-4\}$. To ensure consistency in comparison, all baselines and BioMamba are configured with the same number of blocks $\{6\}$, a batch size of $\{32,32,32,64,128,128\}$, and a hidden dimension of $\{128\}$ across the six tasks. We perform five runs with random seeds $\{2025-2029\}$ and report the mean and standard deviation of the model's performance on the same device. The hyperparameters of BioMamba are listed in Table~\ref{tab: hyperparameters}.

\begin{table*}[htpb]
\centering
\resizebox{\textwidth}{!}{%
\begin{tabular}{cccccccc|cc}
\toprule 
Datasets & Sparsity & Accuracy & Precision & Recall & F1 score & AUROC & AUPRC & Params (M) & FLOPs (G)  \\
\hline 
\multirow{4}{*}{\begin{tabular}{l}
APAVA \\
\end{tabular}} 
& \textbf{ 0.3}  & \deepreds{$84.95 _{\pm 1.35}$} & \deepreds{$85.72 _{\pm 1.95}$} & \deepreds{$83.15 _{\pm 1.13}$} & \deepreds{$83.95 _{\pm 1.32}$} & \deepreds{$93.79 _{\pm 1.39}$} & \deepreds{$93.52 _{\pm 1.40}$} & \deepreds{$0.97$M} & \deepreds{$1.61$G} \\

& 0.5  & $84.84 _{\pm 2.38}$ & $85.38 _{\pm 2.38}$ & $83.16 _{\pm 2.65}$ & $83.86 _{\pm 2.61}$ & $93.68 _{\pm 1.80}$ & $93.28 _{\pm 2.08}$ & $0.89$M & $1.61$G \\
& 0.7  & $84.60 _{\pm 3.40}$ & $84.84 _{\pm 3.83}$ & $83.13 _{\pm 3.40}$ & $83.73 _{\pm 3.53}$ & $93.06 _{\pm 2.96}$ & $92.48 _{\pm 3.49}$ & $0.82$M & $1.61$G \\
& 0.9  & $83.77 _{\pm 1.59}$ & $84.99 _{\pm 1.56}$ & $81.64 _{\pm 2.09}$ & $82.50 _{\pm 1.91}$ & $93.67 _{\pm 1.48}$ & $93.37 _{\pm 1.53}$ & $0.74$M & $1.61$G \\
&  w/o sparsity  & $82.29 _{\pm 3.08}$ & $83.13 _{\pm 2.89}$ & $80.17 _{\pm 3.65}$ & $80.92 _{\pm 3.56}$ & $92.19 _{\pm 2.06}$ & $91.54 _{\pm 2.19}$ & $1.61$M & $1.61$G \\
\hline
\multirow{4}{*}{\begin{tabular}{l}
TDBrain \\
\end{tabular}} 
& 0.3  & $95.94 _{\pm 2.43}$ & $96.06 _{\pm 2.35}$ & $95.94 _{\pm 2.43}$ & $95.93 _{\pm 2.43}$ & $99.18 _{\pm 0.87}$ & $99.19 _{\pm 0.84}$ & $0.98$M & $2.22$G \\
& 0.5  & $96.35 _{\pm 2.62}$ & $96.38 _{\pm 2.59}$ & $96.35 _{\pm 2.62}$ & $96.35 _{\pm 2.62}$ & $99.40 _{\pm 0.60}$ & $99.41 _{\pm 0.59}$ & $0.90$M & $2.22$G \\
& \textbf{ 0.7}  & \deepreds{$96.77 _{\pm 1.94}$} & \deepreds{$96.90 _{\pm 1.71}$} & \deepreds{$96.77 _{\pm 1.94}$} & \deepreds{$96.77 _{\pm 1.95}$} & \deepreds{$99.44 _{\pm 0.49}$} & \deepreds{$99.42 _{\pm 0.51}$} & \deepreds{$0.83$M} & \deepreds{$2.22$G} \\

& 0.9  & $95.65 _{\pm 2.92}$ & $95.74 _{\pm 2.80}$ & $95.65 _{\pm 2.92}$ & $95.64 _{\pm 2.93}$ & $99.22 _{\pm 0.76}$ & $99.24 _{\pm 0.74}$ & $0.75$M & $2.22$G \\
& w/o sparsity  & $95.06 _{\pm 2.72}$ & $95.09 _{\pm 2.70}$ & $95.06 _{\pm 2.72}$ & $95.06 _{\pm 2.72}$ & $98.93 _{\pm 0.95}$ & $98.94 _{\pm 0.90}$ & $1.10$M & $2.22$G \\
\hline
\multirow{4}{*}{\begin{tabular}{l}
Crowdsourced \\
\end{tabular}} 
& 0.3  & $89.07 _{\pm 1.19}$ & $89.37 _{\pm 0.87}$ & $89.06 _{\pm 1.19}$ & $89.04 _{\pm 1.22}$ & $96.74 _{\pm 0.32}$ & $96.82 _{\pm 0.33}$ & $0.97$M & $1.40$G \\
& 0.5  & $89.36 _{\pm 0.81}$ & $89.55 _{\pm 0.76}$ & $89.36 _{\pm 0.81}$ & $89.35 _{\pm 0.82}$ & $96.70 _{\pm 0.37}$ & $96.79 _{\pm 0.37}$ & $0.89$M & $1.40$G \\
& \textbf{0.7}  & \deepreds{$89.84 _{\pm 0.72}$} & \deepreds{$90.04 _{\pm 0.75}$} & \deepreds{$89.83 _{\pm 0.72}$} & \deepreds{$89.82 _{\pm 0.71}$} & \deepreds{$96.88 _{\pm 0.34}$} & \deepreds{$96.97 _{\pm 0.33}$} & \deepreds{$0.81$M} & \deepreds{$1.40$G} \\

& 0.9  & $89.55 _{\pm 1.34}$ & $89.61 _{\pm 1.27}$ & $89.55 _{\pm 1.34}$ & $89.55 _{\pm 1.35}$ & $96.35 _{\pm 0.36}$ & $96.40 _{\pm 0.35}$ & $0.73$M & $1.40$G \\
& w/o sparsity  & $89.42 _{\pm 1.39}$ & $89.53 _{\pm 1.37}$ & $89.42 _{\pm 1.39}$ & $89.42 _{\pm 1.39}$ & $96.59 _{\pm 0.88}$ & $96.63 _{\pm 0.92}$ & $1.08$M & $1.40$G \\
\hline
\multirow{4}{*}{\begin{tabular}{l}
STEW \\
\end{tabular}} 
& 0.3  & $79.30 _{\pm 0.67}$ & $79.31 _{\pm 0.66}$ & $79.30 _{\pm 0.67}$ & $79.30 _{\pm 0.67}$ & $87.08 _{\pm 0.60}$ & $86.83 _{\pm 0.55}$ & $0.97 $M & $1.88$G \\
& 0.5  & $79.13 _{\pm 0.91}$ & $79.29 _{\pm 0.96}$ & $79.13 _{\pm 0.91}$ & $79.11 _{\pm 0.91}$ & $86.98 _{\pm 0.78}$ & $86.70 _{\pm 0.78}$ & $0.89 $M & $1.88$G \\
& 0.7  & $79.27 _{\pm 0.47}$ & $79.34 _{\pm 0.42}$ & $79.27 _{\pm 0.47}$ & $79.26 _{\pm 0.48}$ & $87.08 _{\pm 0.53}$ & $86.80 _{\pm 0.57}$ & $0.81 $M & $1.88$G \\
& \textbf{0.9}  & \deepreds{$79.60 _{\pm 1.00}$} & \deepreds{$79.65 _{\pm 1.03}$} & \deepreds{$79.60 _{\pm 1.00}$} & \deepreds{$79.59 _{\pm 0.99}$} & \deepreds{$87.44 _{\pm 0.56}$} & \deepreds{$87.27 _{\pm 0.53}$} & \deepreds{$0.73$M} & \deepreds{$1.88$G} \\
&  w/o sparsity  & $79.49 _{\pm 0.77}$ & $79.59 _{\pm 0.80}$ & $79.49 _{\pm 0.77}$ & $79.47 _{\pm 0.76}$ & $87.25 _{\pm 0.59}$ & $86.99 _{\pm 0.57}$ & $1.09 $M & $1.88$G \\
\hline
\multirow{4}{*}{\begin{tabular}{l}
DREAMER \\
\end{tabular}} 
& \textbf{ 0.3}  &  {$52.94 _{\pm 3.27}$} & \deepreds{$50.79 _{\pm 2.63}$} & \deepreds{$50.70 _{\pm 2.62}$} & \deepreds{$50.60 _{\pm 2.59}$} & \deepreds{$49.51 _{\pm 4.57}$} & \deepreds{$50.84 _{\pm 3.90}$} & \deepreds{$0.97$M} & \deepreds{$3.76$G} \\

& 0.5  & \deepreds{$53.46 _{\pm 6.15}$} & $50.59 _{\pm 5.87}$ & $50.31 _{\pm 5.44}$ & $49.92 _{\pm 5.41}$ & $48.80 _{\pm 5.33}$ & $50.61 _{\pm 3.72}$ & $0.89$M & $3.76$G \\
& 0.7  & $48.79 _{\pm 4.95}$ & $45.90 _{\pm 4.57}$ & $45.92 _{\pm 4.37}$ & $45.73 _{\pm 4.20}$ & $45.71 _{\pm 5.76}$ & $49.29 _{\pm 4.13}$ & $0.81$M & $3.76$G \\
& 0.9  & $53.09 _{\pm 5.71}$ & $49.95 _{\pm 5.91}$ & $49.81 _{\pm 5.57}$ & $49.38 _{\pm 5.49}$ & $45.86 _{\pm 4.47}$ & $48.80 _{\pm 3.51}$ & $0.73$M & $3.76$G \\
&  w/o sparsity  & $50.38 _{\pm 2.82}$ & $47.59 _{\pm 2.91}$ & $47.69 _{\pm 2.78}$ & $47.51 _{\pm 2.86}$ & $47.36 _{\pm 3.70}$ & $50.12 _{\pm 4.03}$ & $1.09$M & $3.76$G \\
\hline
\multirow{4}{*}{\begin{tabular}{l}
PTB \\
\end{tabular}} 
& 0.3  & \deepreds{$84.63 _{\pm 0.86}$} & $87.49 _{\pm 1.51}$ & $78.06 _{\pm 1.24}$ & \deepreds{$80.53 _{\pm 1.22}$} & $95.00 _{\pm 0.93}$ & $94.12 _{\pm 0.90}$ & $0.98$M & $4.04$G  \\
& 0.5  & $82.98 _{\pm 3.36}$ & $86.52 _{\pm 3.23}$ & $75.43 _{\pm 4.90}$ & $77.76 _{\pm 5.17}$ & $94.42 _{\pm 0.75}$ & $93.35 _{\pm 1.08}$ & $0.90$M & $4.04$G  \\
& \textbf{0.7}  & {$84.53 _{\pm 3.12}$} & \deepreds{$87.50 _{\pm 2.20}$} & \deepreds{$77.86 _{\pm 4.88}$} & {$80.18 _{\pm 4.85}$} & \deepreds{$95.14 _{\pm 0.61}$} & \deepreds{$94.30 _{\pm 1.10}$} & \deepreds{$0.82$M} & \deepreds{$4.04$G}  \\

& 0.9  &$84.13 _{\pm 1.84}$& $87.15 _{\pm 1.38}$ & $77.29 _{\pm 2.94}$ & $79.71 _{\pm 2.84}$ & $94.66 _{\pm 0.67}$ & $93.65 _{\pm 0.58}$ & $0.74$M & $4.04$G  \\
& w/o sparsity  & $81.72 _{\pm 3.75}$ & $85.41 _{\pm 3.10}$ & $73.58 _{\pm 5.56}$ & $75.71 _{\pm 6.06}$ & $94.29 _{\pm 1.78}$ & $93.02 _{\pm 1.99}$ & $1.10$M & $4.04$G  \\
\bottomrule
\end{tabular}
}
\caption{Ablation study on sparsity levels of Feed
Forward module across different datasets. We evaluate sparsity levels $s \in \{0.3, 0.5, 0.7, 0.9\}$ and also compare them to the configuration \textbf{without (w/o)} sparsity. We can observe that the sparsity strategy not only reduces the model's parameters but also leads to better classification performance. The best results are in \textbf{bold}. }
\label{tab: ablation_of_sparsity}
\end{table*}


\begin{table*}[htpb]
\centering
\resizebox{\textwidth}{!}{%
\begin{tabular}{cccccccc|cc}
\toprule
Datasets & Frequency Resolution & Accuracy & Precision & Recall & F1 score & AUROC & AUPRC & Params (M) & FLOPs (G)  \\
\hline 
\multirow{5}{*}{\begin{tabular}{l}
APAVA \\
\end{tabular}} 
& {[256, ~~50]}  & {$83.05 _{\pm 3.65}$} & $83.42 _{\pm 3.43}$ & $81.36 _{\pm 4.38}$ & $81.92 _{\pm 4.24}$ & $92.10 _{\pm 3.05}$ & $91.49 _{\pm 3.39}$ & $0.74$M & $1.08$G \\
& [200, 100] & $81.90 _{\pm 2.25}$ & $83.79 _{\pm 2.80}$ & $79.17 _{\pm 2.31}$ & $80.19 _{\pm 2.46}$ & $92.17 _{\pm 1.27}$ & $91.79 _{\pm 1.11}$ & $0.73$M & $1.07$G \\
& \textbf{[200, ~~50]} & \deepreds{$84.61 _{\pm 3.41}$} & \deepreds{$84.87 _{\pm 3.85}$} & \deepreds{$83.14 _{\pm 3.40}$} & \deepreds{$83.74 _{\pm 3.54}$} & \deepreds{$93.09 _{\pm 2.98}$} & \deepreds{$92.51 _{\pm 3.51}$} & \deepreds{$0.74$M} & \deepreds{$1.61$G} \\

& [128, 100]  & $79.32 _{\pm 2.06}$ & $80.85 _{\pm 1.62}$ & $76.46 _{\pm 2.71}$ & $77.26 _{\pm 2.70}$ & $89.46 _{\pm 1.53}$ & $89.07 _{\pm 1.59}$ & $0.73$M & $1.60$G \\
& [128, ~~50]  & $77.71 _{\pm 2.18}$ & $79.71 _{\pm 2.92}$ & $74.40 _{\pm 2.31}$ & $75.20 _{\pm 2.49}$ & $87.93 _{\pm 4.06}$ & $87.51 _{\pm 3.77}$ & $0.74$M & $2.13$G \\

\hline
\multirow{5}{*}{\begin{tabular}{l}
TDBrain \\
\end{tabular}} 
& \textbf{[256, ~~50]}  & \deepreds{$96.77 _{\pm 1.94}$} & \deepreds{$96.90 _{\pm 1.71}$} & \deepreds{$96.77 _{\pm 1.94}$} & \deepreds{$96.77 _{\pm 1.95}$} & \deepreds{$99.44 _{\pm 0.49}$} & \deepreds{$99.42 _{\pm 0.51}$} & \deepreds{$0.98$M} & \deepreds{$2.22$G} \\

& [200, 100] & $95.77 _{\pm 1.89}$ & $95.83 _{\pm 1.91}$ & $95.77 _{\pm 1.89}$ & $95.77 _{\pm 1.89}$ & $99.07 _{\pm 0.61}$ & $99.01 _{\pm 0.68}$ & $0.98$M & $2.21$G \\
&[200, ~~50] & $94.02 _{\pm 3.24}$ & $94.11 _{\pm 3.23}$ & $94.02 _{\pm 3.24}$ & $94.02 _{\pm 3.24}$ &$98.72 _{\pm 1.17}$ & $98.77 _{\pm 1.10}$ & $0.99$M & $3.31$G \\
& [128, 100]  & $93.52 _{\pm 3.41}$ & $93.74 _{\pm 3.19}$ & $93.52 _{\pm 3.41}$ & $93.51 _{\pm 3.43}$ & $98.49 _{\pm 0.96}$ & $98.53 _{\pm 0.94}$ & $0.98$M & $3.30$G \\
& [128, ~~50]  & $91.83 _{\pm 3.36}$ & $91.91 _{\pm 3.33}$ & $91.83 _{\pm 3.36}$ & $91.83 _{\pm 3.36}$ & $97.72 _{\pm 1.26}$ & $97.79 _{\pm 1.21}$ & $0.99$M & $4.39$G \\
\hline
\multirow{5}{*}{\begin{tabular}{l}
Crowdsourced \\
\end{tabular}} 
& [256, ~~50]  & $88.01 _{\pm 1.46}$ & $88.34 _{\pm 1.18}$ & $88.01 _{\pm 1.46}$ & $87.98 _{\pm 1.49}$ & $96.11 _{\pm 0.54}$ & $96.22 _{\pm 0.50}$ & $0.73$M & $0.94$G \\
& [200, 100] & {$88.96 _{\pm 0.77}$} & $89.00 _{\pm 0.76}$ & $88.96 _{\pm 0.77}$ & $88.96 _{\pm 0.77}$ & $96.27 _{\pm 0.49}$ & $96.39 _{\pm 0.47}$ & $0.73$M & $0.94$G \\
& [200, ~~50] & $89.67 _{\pm 0.52}$ & $89.76 _{\pm 0.44}$ & $89.67 _{\pm 0.52}$ & $89.66 _{\pm 0.52}$ & $96.45 _{\pm 0.34}$ & $96.52 _{\pm 0.38}$ &  $0.73$M & $1.40$G \\
& \textbf{[128, 100]} & \deepreds{$89.84 _{\pm 0.72}$} & \deepreds{$90.04 _{\pm 0.75}$} & \deepreds{$89.83 _{\pm 0.72}$} & \deepreds{$89.82 _{\pm 0.71}$} & \deepreds{$96.88 _{\pm 0.34}$} & \deepreds{$96.97 _{\pm 0.33}$} & \deepreds{$0.73$M} & \deepreds{$1.40$G} \\

& [128, ~~50]  & $88.60 _{\pm 1.38}$ & $88.92 _{\pm 1.04}$ & $88.60 _{\pm 1.38}$ & $88.57 _{\pm 1.42}$ & $96.35 _{\pm 0.43}$ & $96.43 _{\pm 0.42}$ & $0.73$M & $1.86$G \\
\hline
\multirow{5}{*}{\begin{tabular}{l}
STEW \\
\end{tabular}} 
&\textbf{[256, ~~50]}  & \deepreds{$79.27 _{\pm 0.47}$} & \deepreds{$79.34 _{\pm 0.42}$} & \deepreds{$79.27 _{\pm 0.47}$} & \deepreds{$79.26 _{\pm 0.48}$} & \deepreds{$87.08 _{\pm 0.53}$} & {$86.80 _{\pm 0.57}$} & \deepreds{$0.73$M} & \deepreds{$1.88$G} \\

& [200, 100] & $78.27 _{\pm 0.49}$ & $78.31 _{\pm 0.50}$ & $78.27 _{\pm 0.49}$ & $78.26 _{\pm 0.49}$ & $86.17 _{\pm 0.27}$ & $85.91 _{\pm 0.29}$ & $0.73$M & $1.88$G \\
& [200, ~~50] & $78.47 _{\pm 1.09}$ & $78.51 _{\pm 1.09}$ & $78.47 _{\pm 1.09}$ & $78.46 _{\pm 1.09}$ & $87.00 _{\pm 0.99}$ & \deepreds{$86.91 _{\pm 0.93}$} & $0.73$M & $2.81$G \\
& [128, 100]  & $78.63 _{\pm 0.55}$ & $78.67 _{\pm 0.52}$ & $78.63 _{\pm 0.55}$ & $78.62 _{\pm 0.55}$ & $86.91 _{\pm 0.33}$ & $86.74 _{\pm 0.36}$ & $0.73$M & $2.80$G \\
& [128, ~~50]  & $78.40 _{\pm 0.63}$ & $78.49 _{\pm 0.70}$ & $78.40 _{\pm 0.63}$ & $78.38 _{\pm 0.62}$ & $86.83 _{\pm 0.39}$ & $86.72 _{\pm 0.36}$ & $0.73$M & $3.73$G \\
\hline
\multirow{5}{*}{\begin{tabular}{l}
DREAMER \\
\end{tabular}} 
& \textbf{[256, ~~50]}  & {$52.71 _{\pm 3.25}$} & {$50.51 _{\pm 2.54}$} & \deepreds{$50.42 _{\pm 2.51}$} & \deepreds{$50.32 _{\pm 2.46}$} & \deepreds{$49.42 _{\pm 4.69}$} & \deepreds{$50.83 _{\pm 3.92}$} & \deepreds{$1.09$M} & \deepreds{$3.76$G} \\

& [200, 100] & $50.69 _{\pm 2.32}$ & $47.03 _{\pm 2.40}$ & $47.29 _{\pm 2.19}$ &$46.84 _{\pm 2.31}$ & $43.86 _{\pm 4.05}$ & $46.82 _{\pm 2.71}$ & $1.09$M & $3.76$G \\
& [200, ~~50] & $49.92 _{\pm 7.07}$ & $45.62 _{\pm 8.57}$ & $46.12 _{\pm 7.28}$ & $45.40 _{\pm 7.74}$ & $42.08 _{\pm 5.16}$ & $46.37 _{\pm 3.32}$ & $1.09$M & $5.62$G \\
& [128, 100]  & \deepreds{$53.88 _{\pm 4.82}$} & \deepreds{$50.78 _{\pm 4.68}$} & $50.30 _{\pm 3.74}$ & $49.73 _{\pm 3.31}$ & $46.68 _{\pm 1.37}$ & $49.15 _{\pm 1.05}$ & $1.09$M & $5.60$G \\
& [128, ~~50]  & $49.63 _{\pm 6.46}$ & $45.46 _{\pm 6.76}$ & $45.60 _{\pm 5.48}$ & $44.57 _{\pm 5.02}$ & $43.98 _{\pm 2.98}$ & $48.98 _{\pm 2.55}$ & $1.09$M & $7.45$G \\
\hline
\multirow{5}{*}{\begin{tabular}{l}
PTB \\
\end{tabular}} 
& \textbf{[256, ~~50]} & \deepreds{$84.53 _{\pm 3.12}$} & \deepreds{$87.50 _{\pm 2.20}$} & \deepreds{$77.86 _{\pm 4.88}$} & \deepreds{$80.18 _{\pm 4.85}$} & {$95.14 _{\pm 0.61}$} & {$94.30 _{\pm 1.10}$} & \deepreds{$0.98$M} & \deepreds{$4.04$G} \\

& [200, 100] & $81.78 _{\pm 3.20}$ & $86.08 _{\pm 2.17}$ & $73.46 _{\pm 4.84}$ & $75.68 _{\pm 5.24}$ & $94.97 _{\pm 1.40}$ & $93.92 _{\pm 1.39}$ & $0.98$M & $6.03$G \\
& [200, ~~50] & $83.02 _{\pm 3.33}$ & $86.69 _{\pm 1.36}$ & $75.55 _{\pm 5.51}$ & $77.73 _{\pm 5.59}$ & $93.67 _{\pm 1.86}$ & $92.77 _{\pm 1.59}$ & $0.98$M & $8.02$G \\
& [128, 100]  & $83.05 _{\pm 3.69}$ & $86.80 _{\pm 2.74}$ & $75.42 _{\pm 5.49}$ & $77.74 _{\pm 5.66}$ & {$96.42 _{\pm 0.52}$} & {$95.24 _{\pm 0.77}$} & $0.98$M & $6.01$G \\
& [128, ~~50]  & $82.75 _{\pm 1.62}$ & $87.26 _{\pm 1.97}$ & $74.75 _{\pm 2.24}$ & $77.29 _{\pm 2.43}$ & \deepreds{$96.95 _{\pm 0.68}$} & \deepreds{$95.98 _{\pm 0.96}$} & $0.98$M & $9.98$G \\
\bottomrule
\end{tabular}
}
\caption{Ablation study on varying frequency resolutions across different datasets. We use frequency pairs \{[256, 50], [200, 100], [200, 50], [128, 100], [128, 50]\} for analysis. The best results are in \textbf{bold}.}
\label{tab: ablation_of_frequency_resolution}
\end{table*}

\begin{table*}[htpb]
\centering
\resizebox{\textwidth}{!}{%
\begin{tabular}{cccccccc|cc}
\toprule Datasets & Mamba & Accuracy & Precision & Recall & F1 score & AUROC & AUPRC & Params (M) & FLOPs (G)  \\
\hline 
\multirow{2}{*}{\begin{tabular}{l}
APAVA \\
\end{tabular}} 
& w/o Bidirectional  & \deepreds{$85.09 _{\pm 2.31}$} & $85.40 _{\pm 2.67}$ & \deepreds{$83.62 _{\pm 2.29}$} & \deepreds{$84.23 _{\pm 2.38}$} & $93.68 _{\pm 1.75}$ & $93.22 _{\pm 2.17}$ & $0.66$M & $1.13$G \\
& BioMamba & {$84.95 _{\pm 1.35}$} & \deepreds{$85.72 _{\pm 1.95}$} & {$83.15 _{\pm 1.13}$} & {$83.95 _{\pm 1.32}$} & \deepreds{$93.79 _{\pm 1.39}$} & \deepreds{$93.52 _{\pm 1.40}$} & \deepreds{$0.97$M} & \deepreds{$1.61$G} \\
\hline
\multirow{2}{*}{\begin{tabular}{l}
TDBrain \\
\end{tabular}} 
& w/o Bidirectional  & $96.12 _{\pm 1.61}$ & $96.22 _{\pm 1.45}$ & $96.12 _{\pm 1.61}$ & $96.12 _{\pm 1.62}$ & \deepreds{$99.52 _{\pm 0.14}$} & \deepreds{$99.54 _{\pm 0.14}$} & $0.51$M & $1.56$G \\
& BioMamba & \deepreds{$96.77 _{\pm 1.94}$} & \deepreds{$96.90 _{\pm 1.71}$} & \deepreds{$96.77 _{\pm 1.94}$} & \deepreds{$96.77 _{\pm 1.95}$} & {$99.44 _{\pm 0.49}$} & {$99.42 _{\pm 0.51}$} & \deepreds{$0.83$M} & \deepreds{$2.22$G} \\
\hline
\multirow{2}{*}{\begin{tabular}{l}
Crowdsourced \\
\end{tabular}} 
& w/o Bidirectional  & $89.62 _{\pm 1.00}$ & $89.95 _{\pm 0.87}$ & $89.62 _{\pm 1.00}$ & $89.60 _{\pm 1.01}$ & \deepreds{$97.03 _{\pm 0.38}$} & \deepreds{$97.10 _{\pm 0.37}$} & $0.49$M & $0.98$G \\
& BioMamba & \deepreds{$89.84 _{\pm 0.72}$} & \deepreds{$90.04 _{\pm 0.75}$} & \deepreds{$89.83 _{\pm 0.72}$} & \deepreds{$89.82 _{\pm 0.71}$} & {$96.88 _{\pm 0.34}$} & {$96.97 _{\pm 0.33}$} & \deepreds{$0.81$M} & \deepreds{$1.40$G} \\
\hline
\multirow{2}{*}{\begin{tabular}{l}
STEW \\
\end{tabular}} 
& w/o Bidirectional  & $79.08 _{\pm 0.66}$ & $79.17 _{\pm 0.70}$ & $79.08 _{\pm 0.66}$ & $79.06 _{\pm 0.65}$& $87.09 _{\pm 0.66}$ & $86.86 _{\pm 0.68}$ & $0.42$M & $1.32$G \\
& BioMamba & \deepreds{$79.60 _{\pm 1.00}$} & \deepreds{$79.65 _{\pm 1.03}$} & \deepreds{$79.60 _{\pm 1.00}$} & \deepreds{$79.59 _{\pm 0.99}$} & \deepreds{$87.44 _{\pm 0.56}$} & \deepreds{$87.27 _{\pm 0.53}$} & \deepreds{$0.73 $M} & \deepreds{$1.88$G} \\
\hline
\multirow{2}{*}{\begin{tabular}{l}
DREAMER \\
\end{tabular}} 
& w/o Bidirectional  & \deepreds{$53.20 _{\pm 4.29}$} & \deepreds{$51.16 _{\pm 3.89}$} & \deepreds{$51.10 _{\pm 3.92}$} & \deepreds{$51.00 _{\pm 3.94}$} & $48.09 _{\pm 3.87}$ & $50.01 _{\pm 3.61}$ & $0.66$M & $2.65$G \\
& BioMamba & {$52.94 _{\pm 3.27}$} & {$50.79 _{\pm 2.63}$} & {$50.70 _{\pm 2.61}$} & {$50.60 _{\pm 2.58}$} & \deepreds{$49.51 _{\pm 4.57}$} & \deepreds{$50.84 _{\pm 3.90}$} & \deepreds{$0.97$M} & \deepreds{$3.76$G} \\
\hline
\multirow{2}{*}{\begin{tabular}{l}
PTB \\
\end{tabular}} 
& w/o Bidirectional  & $84.15 _{\pm 2.58}$ & $87.47 _{\pm 1.92}$ & $77.19 _{\pm 4.05}$ & $79.59 _{\pm 4.15}$ & \deepreds{$95.35 _{\pm 1.06}$} & \deepreds{$94.54 _{\pm 0.74}$ }& $0.50$M & $2.85$G \\
& BioMamba & \deepreds{$84.53 _{\pm 3.12}$} & \deepreds{$87.50 _{\pm 2.20}$} & \deepreds{$77.86 _{\pm 4.88}$} & \deepreds{$80.18 _{\pm 4.85}$} & {$95.14 _{\pm 0.61}$} & {$94.30 _{\pm 1.10}$} & \deepreds{$0.82$M} & \deepreds{$4.04$G}  \\
\bottomrule
\end{tabular}
}
\caption{Ablation study on different Mamba configurations to analyze the impact of Bidirectional Mamba. Configurations without (w/o) Bidirectional Mamba are compared to the default model. The best results are in \textbf{bold}.}
\label{tab: ablation_of_bidmamba}
\end{table*}
\section{Ablation Studies}
\label{sec: Ablation Studies}

We perform comprehensive ablation studies on the key components and hyperparameter choices of BioMamba, reporting performance across six datasets.
These studies help highlight the impact of each component on model effectiveness and provide insights into optimal configurations for biosignal analysis.

\textbf{Embedding Types.}
Table~\ref{tab: ablation_of_embedding} shows the effects of Patched Spectral Embedding (PSE) and Temporal Domain Embedding (TDE). Specifically, removing the PSE component leads to a notable reduction in performance, attributed to the spectral magnitude information it provides, which complements the temporal domain information in biosignals. 
This demonstrates the superior
performance of our proposed embedding approach for the Bidirectional Mamba model learning. Notably, on the STEW dataset, the PSE approach outperforms the Spectro-Temporal Embedding strategy, while showing higher variance across four evaluation metrics, due to the loss of temporal information introducing instabilities in the TDE component. Figure~\ref{fig: crowdsource_dataset} shows the frequency and temporal information of the binary classes in the Crowdsource datasets, which also explains why the Spectro-Temporal Embedding strategy works for biosignal classification.

\textbf{Sparse Feed Forward.}
Table~\ref{tab: ablation_of_sparsity} presents the various sparsity levels of the BioMamba blocks. We set the Frequency Resolution as the default configuration, as shown in Table~\ref{tab: hyperparameters}. Based on Table~\ref{tab: ablation_of_sparsity}, different sparsity levels require varying computational resources and affect performance; however, all sparsity levels yield only minor differences in results. As observed, the performance gap among different sparsity levels is negligible, while the difference in performance and computational efficiency between the Sparse Feed Forward and non-sparse configurations is significant. For instance, In the PTB dataset, applying a sparsity level of 0.7 achieves a precision of 87.5\%. In contrast, without sparsity, performance decreases by 2.09\%, and the number of model parameters is reduced from 1.10M to 0.82M. This ablation study effectively highlights the benefits of sparsity regarding computational efficiency and performance.

\textbf{Frequency Resolution.}
We provide ablation study on six different frequency resolutions $\{a,b\}$ in Table~\ref{tab: ablation_of_frequency_resolution} to evaluate the effect of frequency bins $\{a\}$ and window shifts $\{b\}$. We find that the larger frequency resolution achieves the highest performance across all evaluated metrics.

\textbf{Bidirectional Mamba.} To evaluate the effect of different Mamba configurations, we evaluate both the Unidirectional and the Bidirectional Mamba block for the BioMamba. The results are shown in Table~\ref{tab: ablation_of_bidmamba}. The results indicate that the Bidirectional Mamba block achieves a better performance on most datasets than the unidirectional structure.

\begin{figure*}[htbp]
\centering
\includegraphics[width=1\textwidth]{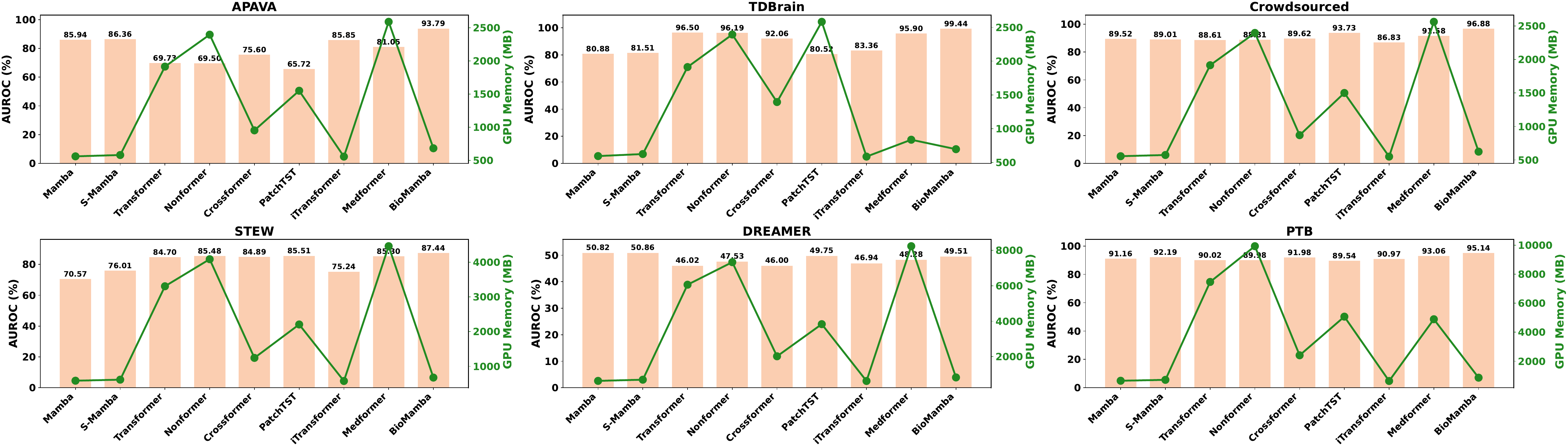}
\caption{AUROC and GPU memory utilization analysis, BioMamba outperforms previous models in biosignal classification across five datasets with the least GPU memory usage. The numerical results are listed in Table~\ref{tab: model_efficiency}.}
\label{fig: gpu_memory}
\end{figure*}

\begin{figure*}[htbp]
\centering
\includegraphics[width=1\textwidth]{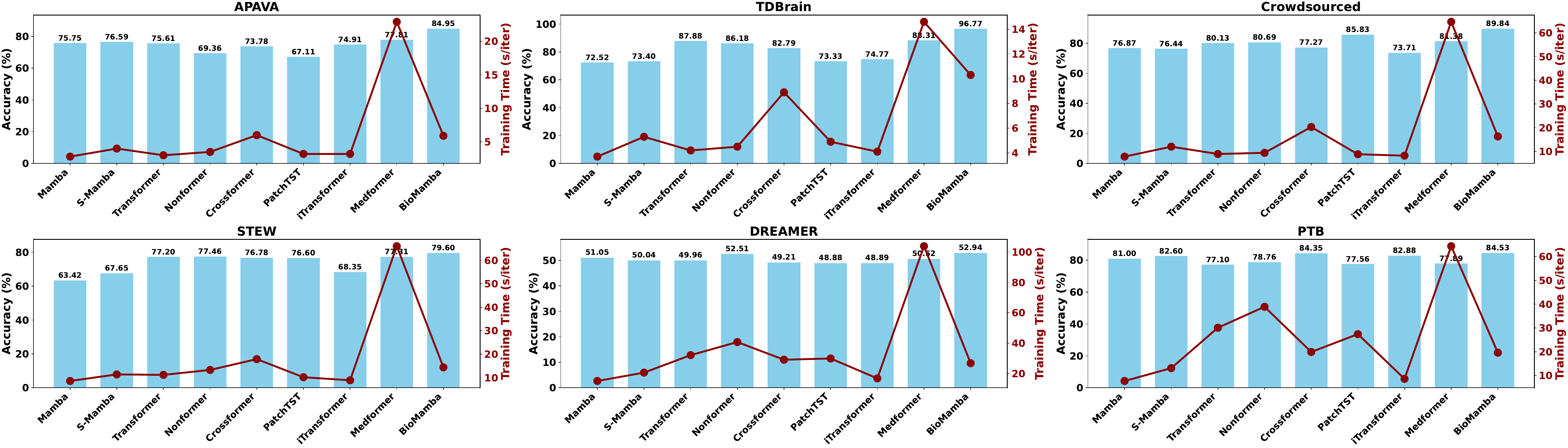}
\caption{Accuracy and training time analysis, BioMamba represents the best classification accuracy across all methods with comparable training efficiency in six datasets. The numerical results are listed in Table~\ref{tab: model_efficiency}.}
\label{fig: training_time}
\end{figure*}

\section{Training Efficiency, Average Experimental Results, and Visualization}
\subsection{Training Efficiency Comparison} 

To provide a clearer assessment of BioMamba's efficiency, we evaluate the training time per epoch and GPU memory consumption for each dataset, as shown in Figure~\ref{fig: gpu_memory} and Figure~\ref{fig: training_time}. According to the table, BioMamba achieves $1\times$-$10\times$ improvement of Medformer in GPU memory consumption across the six different tasks. Notably, Additionally, we observe that Medformer~\cite{wang2024medformer} demonstrates inefficient learning outcomes, as reflected in its high training time and GPU memory usage.

\begin{table*}[htpb]
\centering
\resizebox{\textwidth}{!}{%
\begin{tabular}{l|cc|cc|cc|cc|cc|cc}
\toprule
{Datasets}  & \multicolumn{2}{c|}{APAVA} & \multicolumn{2}{c|}{TDBrain} & \multicolumn{2}{c|}{Crowdsourced}  & \multicolumn{2}{c}{STEW} & \multicolumn{2}{c|}{DREAMER}  & \multicolumn{2}{c}{PTB} \\

\hline
\diagbox[innerleftsep=0.1cm, innerrightsep=0.2cm]{Models}{Efficiency}   & \makecell{Training \\ Times \\ (s/iter)}    & \makecell{GPU \\ Memory \\ (MB)}  & \makecell{Training \\Times \\ (s/iter)}    & \makecell{GPU \\Memory \\ (MB)}& \makecell{Training \\Times \\ (s/iter)}    & \makecell{GPU \\Memory \\ (MB)}& \makecell{Training \\Times \\ (s/iter)}    & \makecell{GPU \\Memory \\ (MB)}& \makecell{Training \\Times \\ (s/iter)}    & \makecell{GPU \\Memory \\ (MB)}& \makecell{Training \\Times \\ (s/iter)}    & \makecell{GPU \\Memory \\ (MB)} \\
\hline
Mamba        & 2.8 & 560 & 3.7 & 594  & 7.8 & 556 & 8.6 & 584 & 15.1 & 636 & 7.7 & 652 \\
S-Mamba      & 4.0 & 580 & 5.3 & 624 & 12.0 & 574 & 11.4 & 614 & 20.6 & 702 & 13.1 & 714 \\
Transformer  & 3.0 & 1914 & 4.2 & 1912 & 8.9 & 1914 & 11.2 & 3308 & 32.1 & 6058 & 30.1 & 7466 \\
Nonformer    & 3.5 & 2396 & 4.5 & 2394 & 9.4 & 2396 & 13.3 & 4088 & 40.7 & 7328 & 38.9 & 9932 \\
Crossformer  & 6.0 & 952 & 8.9 & 1394 & 20.3 & 870 & 17.9 & 1242 & 29.1 & 2022 & 19.9 & 2408 \\
PatchTST     & 3.2 & 1550 & 4.9 & 2582 & 8.8 & 1500 & 10.2 & 2206 & 29.9 & 3834 & 27.4 & 5072 \\
iTransformer & 3.2 & 556 & 4.1 & 586 & 8.2 & 550 & 8.9 & 576 & 16.8 & 630 & 8.6 & 636 \\
Medformer    & 22.9 & 2590 & 14.6 & 836 & 64.7 & 2562 & 66.0 & 4464 & 103.8 & 8230 & 64.4 & 4894 \\
\hline 
 \rowcolor{purple!20}
BioMamba     & 5.9 & 682 & 10.3 & 696 & 16.3 & 624 & 14.4 & 674 & 26.8 & 830 & 19.6 & 864 \\

Improve. &  \deepred{$ 4\times $}  &  \deepred{$ 4\times $}  &  \deepred{$ 1 \times $}  &  \deepred{$ 1 \times $}  &  \deepred{$ 4 \times $}   &  \deepred{$ 4 \times $}   &  \deepred{$ 5 \times $}  &  \deepred{$ 7 \times $} 
&  \deepred{$ 4 \times $}  &  \deepred{$ 10 \times $}
&  \deepred{$ 3 \times $} &  \deepred{$ 6 \times $}
\\
\bottomrule
\end{tabular}}
\caption{Training efficiency comparison on six datasets, The improvement of BioMamba over the baseline (Medformer~\cite{wang2024mamba}) are in \deepred{red bold}.}
\label{tab: model_efficiency}
\end{table*}

\hypersetup{
    urlcolor=black 
}

\subsection{Average Experimental Results}

We averaged the performance of BioMamba and eight baselines across six datasets, as shown in Table~\ref{tab: average_rank}. From this table, it is evident that our model achieved a~\textbf{5\%-7\%} improvement over Medformer~\cite{wang2024medformer}, establishing a new state-of-the-art result. The code will be released.

\begin{table*}[htpb]
\centering
\resizebox{\textwidth}{!}{%
\begin{tabular}{lcccccc|c}
\toprule
Models & Accuracy & Precision & Recall & F1 score & AUROC & AUPRC& Publication  \\
\hline 
Mamba~\cite{gu2023mamba} & $68.75$ & $70.79$ & $67.8$ & $68.08$ & $78.15$ & $77.80$ &  ArXiv 2023\\
S-Mamba~\cite{wang2024mamba} & $71.12$ & $71.77$ & $69.08$ & $69.41$ & $79.32$ & $79.13$  & ArXiv 2024 \\
Transformer~\cite{vaswani2017attention} & $74.65$ & $75.13$ & $71.93$ & $72.17$ & $79.26$ & $78.96$ & NeurIPS 2017 \\
Nonformer~\cite{liu2022non} & $74.16$ & $74.46$ & $71.72$ & $71.88$ & $79.53$ & $79.08$  & NeurIPS 2022\\
Crossformer~\cite{zhang2023crossformer} & $74.03$ & $75.49$ & $71.75$ & $71.97$ & $80.03$ & $80.31$ & ICLR 2023 \\
PatchTST~\cite{nie2022time} & $71.55$ & $73.54$ & $68.28$ & $67.80$ & $77.46$ & $77.09$ & ICLR 2023 \\
iTransformer~\cite{liu2023itransformer} & $70.59$ & $71.43$ & $68.35$ & $68.63$ & $78.20$ & $78.08$ & ICLR 2024 \\
Medformer~\cite{wang2024medformer} & \deepblue{$75.54$} & \deepblue{$76.54$} & \deepblue{$72.95$} & \deepblue{$73.26$} & \deepblue{$82.53$} & \deepblue{$82.53$}  & NeurIPS 2024\\
\hline
BioMamba & \deepred{$81.44$} & \deepred{$81.77$} & \deepred{$79.65$} & \deepred{$80.15$} & \deepred{$87.03$} & \deepred{$87.05$} &  Ours \\
Improve. & \deepred{$+ 5.90$} & \deepred{$+ 5.23$} & \deepred{$+ 6.70$} & \deepred{$+ 6.89$} & \deepred{$+ 4.50$} & \deepred{$+ 4.52$} & -\\
\bottomrule
\end{tabular}
}
\caption{Average performance of BioMamba and eight baselines across six datasets. The best results for each dataset are in \deepred{red bold}, while baseline (Medformer~\cite{wang2024mamba}) performances are in \deepblue{blue bold}.}
\label{tab: average_rank}
\end{table*}

\begin{figure*}[t]
    \centering
    \includegraphics[width=1\linewidth]{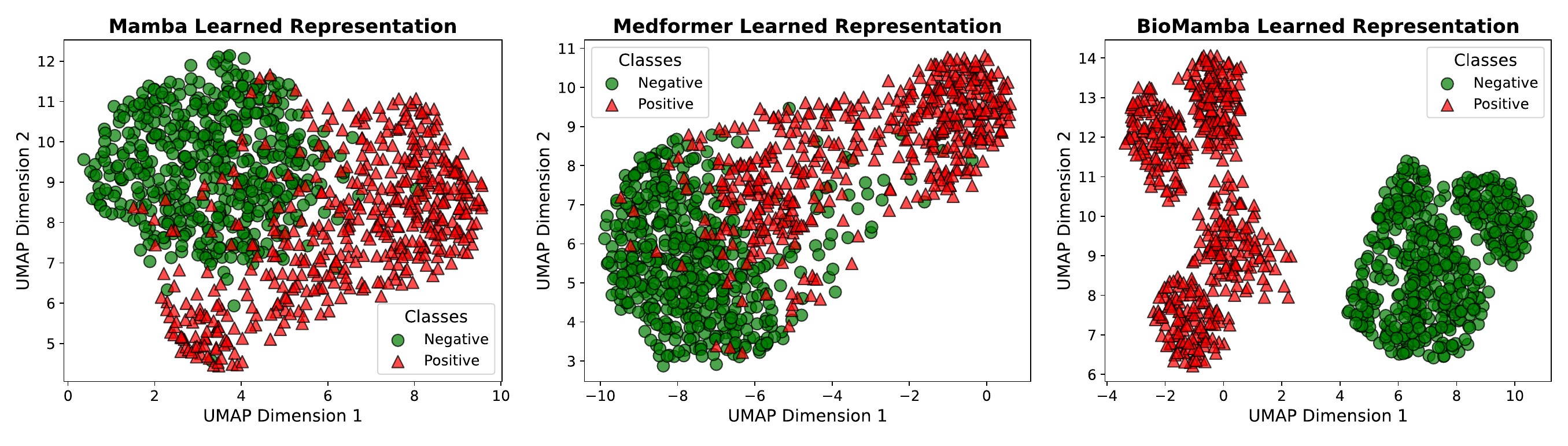} 
    \caption{Visualizing the learned representations from Mamba~\cite{gu2023mamba}, Medformer~\cite{wang2024medformer}, and our BioMamba. The visualized representations were trained from the encoder blocks on the TDBrain dataset~\cite{van2022two}. The green spheres and red triangles represent the negative class (Healthy) and the positive class (Parkinson’s disease), respectively. The results indicate that our approach more effectively segregates the two classes.}
    \label{fig: TD_dataset}
\end{figure*}

\subsection{Visualization}
To visualize the effectiveness of BioMamba, we depict the learned representation $\boldsymbol{Z^{''}}$ from the BioMamba Block, which setup on the TDBrain dataset~\cite{van2022two} as a case study. To visualize the representations more interpretably, we employ UMAP~\cite{mcinnes2018umap}, a dimensionality reduction technique with 50 neighbors and a minimum distance
of 0.5. To establish a reference standard, we utilize Mamba~\cite{gu2023mamba} and Medformer~\cite{wang2024medformer}, since Mamba offers a novel architecture compared to traditional attention-based methods and Medformer shows the best performance among other baselines.

\label{sec: complexity}
\begin{table*}[htpb]
    \centering
    \resizebox{0.5\linewidth}{!}{%
    \begin{tabular}{l|c}
\toprule
    Methods & Computational Complexity \\
\hline
Mamba~\cite{gu2023mamba} & $\mathcal{O}\left(C\right)$ \\
S-Mamba~\cite{wang2024mamba} & $\mathcal{O}\left(C\right)$ \\
Transformer~\cite{vaswani2017attention} & $\mathcal{O}\left(T^2\right)$\\
Nonformer~\cite{liu2022non}  & $\mathcal{O}\left(T^2\right)$ \\
Crossformer~\cite{zhang2023crossformer} & $\mathcal{O}\left(\frac{C T^2}{P^2}\right)$ \\
PatchTST~\cite{nie2022time} & $\mathcal{O}\left(\frac{T^2}{P^2}\right)$ \\
iTransformer~\cite{liu2023itransformer} & $\mathcal{O}\left(C^2\right)$ \\
Medformer~\cite{wang2024medformer} & $\mathcal{O}\left(T^2\right)$\\
\hline
BioMamba (Ours) & $\mathcal{O}\left(\frac{C T}{P}\right)$ \\

\bottomrule
    \end{tabular}
    }
    \caption{Computational complexity analysis.}
    \label{tab: complexity}
\end{table*}

\section{Complexity Analysis}

This section presents a complexity analysis of the proposed eight methods and our BioMamba. As shown in Table~\ref{tab: complexity}, with the variable-wise embedding strategy, the computational complexity of the  Mamba~\cite{gu2023mamba} and S-Mamba~\cite{wang2024mamba} is $\mathcal{O}\left(C\right)$, where $C$ represents the number of channels. The original Transformer~\cite{vaswani2017attention}, Nonformer~\cite{liu2022non}, and Medoformer~\cite{wang2024medformer}, relying on self-attention mechanisms, have time complexity of $\mathcal{O}\left(T^2\right)$, where $T$ denotes the time sequence length.
Crossformer proposed a router mechanism to reduce the complexity to $\mathcal{O}\left(\tfrac{C T^2}{P^2}\right)$, and PatchTST segments time series data into blocks, effectively
distributing the computational to $\mathcal{O}\left(\tfrac{T^2}{P^2}\right)$,  where P denotes the patch size. iTransfomer ~\cite{liu2023itransformer} introduced variable-wise embedding with self-attention mechanism, which presents the complexity with $\mathcal{O}\left(C^2\right)$.
In our BioMamba model, the computational complexity of the patched frequency domain is $\mathcal{O}\left(\frac{C T}{P}\right)$, while that of the temporal domain is $\mathcal{O}(C)$. Consequently, the overall computational complexity of BioMamba remains $\mathcal{O}\left(\frac{C T}{P}\right)$. The computational complexity of BioMamba is significantly lower than that of Medoformer~\cite{wang2024medformer}, specifically $\mathcal{O}\left(\frac{C T}{P}\right) \ll \mathcal{O}\left(T^2\right)$, thereby offering a more efficient solution compared to the quadratic complexity inherent in the attention mechanism of Transformers.

\section{Further Experiments}
\label{sec: ad_experiments}

We further evaluate the performance and efficiency of the BioMamba on four different datasets for multiclass classification tasks, including ADFTD~\cite{miltiadous2023dataset, miltiadous2023dice}, PTB-XL~\cite{wagner2020ptb}, UCI-HAR~\cite{anguita2013public}, and FLAAP~\cite{kumar2022flaap}. And we campare our method with seven approaches: {Mamba}~\cite{gu2023mamba}, {S-Mamba}\cite{wang2024mamba}, TCN~\cite{bai2018empirical}, Transformer~\cite{vaswani2017attention}, {Crossformer}~\cite{zhang2023crossformer},{PatchTST}~\cite{nie2022time}, {Medformer}~\cite{wang2024medformer}. We first provide the details of the datasets and implementation setups, followed by a comparison of classification performance and model efficiency.

\subsection{Datasets}

\begin{table*}[htpb]
\centering
\resizebox{\textwidth}{!}{%
\begin{tabular}{cccccccccccccc}
\toprule
Datasets & Subject & Sample & Class &  Channel  & Timestamps & Sampling Rate & Modality  & Tasks  \\ 
\hline
ADFTD & 88 & 69,752 & 3 & 19 & 256 & 256 Hz & EEG & Brain Diseases Detection \\
PTB-XL & 17,596 & 191,400 & 5 & 12 & 250 & 250 Hz & ECG  & Heart Diseases Classification\\
UCI-HAR & 30 & 10,299 & 6 & 9 & 128 & 50 Hz & Wearable Sensors  &  Human Activity
Recognition\\
FLAAP & 8 & 13,123 & 10 & 6 & 100 & 100 Hz & Wearable Sensors  & Human Activity
Recognition\\

\bottomrule
\end{tabular}}
\caption{Overview of biosignal datasets for further experiments.}
\label{tab: all_datasets_1}
\end{table*}

\textbf{ADFTD}~\cite{miltiadous2023dataset, miltiadous2023dice} is the Alzheimer’s Disease and Frontotemporal Dementia dataset with 3 classes, including 36 Alzheimer's disease (AD) patients, 23 Frontotemporal Dementia (FTD) patients, and 29 healthy control (HC) subjects. The dataset has 19 channels, and the raw sampling rate is 500 Hz. Each subject has a trial, with trial durations of approximately 13.5 minutes for AD subjects ( $\min =5.1$, max $=21.3$ ), 12 minutes for FD subjects $(\min =7.9, \max =16.9$ ), and 13.8 minutes for HC subjects ( $\mathrm{min}=12.5$, $\max =16.5$ ).  Following the Medformer, we set a filter between $0.5-45 \mathrm{~Hz}$ to each trial, downsample each trial to 256 Hz, and segment them into non-overlapping 1-second samples with 256 timestamps, discarding any samples shorter than 1 second. For the subject-independent setup, we set $60 \%, 20 \%$, and $20 \%$ of total subjects with their corresponding samples into the training, validation, and test sets, respectively.

\textbf{PTB-XL}~\cite{wagner2020ptb} is a public ECG dataset recorded from 18,869 subjects, with 12 channels and 5 labels, including Normal ECG, Conduction Disturbance, Myocardial Infarction,  Hypertrophy, ST/T change.  The raw trials consist of 10-second time intervals, with sampling frequencies of 100 Hz and 500 Hz versions. As same as Medformer, we apply the 500 Hz version in 17,596 subjects, then downsample to 250 Hz and normalize. For the training, validation, and test set splits, we allocate $60 \%, 20 \%$, and $20 \%$ of the total subjects for subject-independent learning.

\textbf{UCI-HAR}~\cite{anguita2013public} is a public human activity recognition dataset recorded from the Accelerometer and Gyroscope sensors in a smartphone with 30 subjects and 6 labels, including: Walk,  Walk Upstairs,  Walk Downstairs, Sit, Stand, and Laying. The samples are already split and provided
in the original datasets. 

\textbf{FLAAP}~\cite{kumar2022flaap} (Finding and Learning the Associated Activity Patterns dataset)  is collected from smartphone
accelerometer and gyroscope sensors with  10 labels, the activities including: Sitting, Standing, CrossLeg,  Laying,  Walking,  Jogging,  Cir Walk, StairUp,  StairDown,  SitUp. For the training, validation, and test set splits, we employ the subject-independent setup. Specifically, we allocate $60\%, 20 \%$, and $20\%$ of the total subjects, along with their corresponding samples, into the training, validation, and test sets.

\subsection{Setups}

\begin{table*}[htpb]
\centering
\resizebox{0.6\textwidth}{!}{%
\begin{tabular}{c|cccc}
\toprule
Hyperparameters & ADFTD & PTB-XL & UCI-HAR &  FLAAP   \\
\hline 
Frequency Resolution & [32,~~16] &  [32,~~16] &  [64,~~16] &  [50,~~25]  \\
Sparsity             & 0.3 & 0.3 &  0.3 & 0.3   \\
BioMamba Blocks      & 6 & 6 & 6 & 6   \\
Hidden Dimension     & 128 & 128 & 128  & 128  \\
Batch Size           & 128 & 256 & 32 & 32          \\
Learning Rate  & 5e-5 & 5e-5 &5e-5  &   5e-5        \\
Training Epochs & 100 &  100&  100&100   \\
\bottomrule 
\end{tabular}
}
\caption{Hyperparameters for BioMamba in the further experiments.}
\label{tab: hyperparameters_1}
\end{table*}

We maintain the same hardware and software setups as described in Section~\ref{subsec: setups} for all further experiments. The hyperparameters for BioMamba are listed in Table~\ref{tab: hyperparameters_1}. For the other models, we keep the batch size, blocks, and training epochs identical to those used for BioMamba.


\subsection{Classification Performance and Model Efficiency}

\begin{table*}[htpb]
\centering
\resizebox{\textwidth}{!}{%
\begin{tabular}{c|cc|cc|cc|cc}
\toprule
\multirow{2}{*}{Datasets}  & \multicolumn{2}{c|}{ADFTD} & \multicolumn{2}{c|}{PTB-XL} & \multicolumn{2}{c|}{UCI-HAR}  & \multicolumn{2}{c}{FLAAP} \\
& \multicolumn{2}{c|}{(3-Classes)} & \multicolumn{2}{c|}{(5-Classes)} & \multicolumn{2}{c|}{(6-Classes)}  & \multicolumn{2}{c}{(10-Classes)} \\
\hline
\diagbox[innerleftsep=0.1cm, innerrightsep=0.2cm]{Models}{Efficiency}   & Params (M) &FLOPs (G) &  Params (M) &FLOPs (G) & Params (M) &FLOPs (G) & Params (M) &FLOPs (G)  \\
\hline
Mamba  & 0.76 M & 1.91 G & 0.75 M   & 2.48 G  & 0.74 M  &  0.23 G  & 0.74 M  &  0.16 G \\
S-Mamba  & 1.07 M & 2.71 G & 1.07 M  & 3.52 G &  1.05 M & 0.33 G  & 1.05 M  &  0.23 G \\
TCN & 1.03 M & 33.47 G & 1.02 M  & 65.14 G & 1.02 M  & 4.16 G & 1.02 M & 3.25 G \\
Transformer  & 0.90 M  & 39.18 G & 0.96 M  & 75.78 G & 0.89 M & 4.04 G & 0.92 M &  3.02 G\\
Crossformer & 5.25 M & 31.83 G &  5.23 M & 39.62 G & 5.16 M  & 2.02 G  & 5.15 M & 1.19 G \\
PatchTST & 1.03 M  & 65.87 G & 1.04 M  & 80.48 G & 0.90 M & 3.76 G & 0.88 M  & 1.87 G \\
Medformer  & 8.12 M &  47.83 G & 7.91 M  & 90.11 G & 2.49 M  & 2.75 G & 2.11 M & 2.51 G \\
\hline 
BioMamba  & 1.07 M & 40.07 G & 1.06 M  & 47.46 G & 0.98 M & 1.78 G  & 0.96 M  & 0.79 G  \\
\bottomrule
\end{tabular}
}
\caption{Comparison of model efficiency in the additional experiments.}
\label{tab: performance_2}
\end{table*}

We present the multiclass classification results in Table~\ref{tab: performance_1}. BioMamba achieves a new state-of-the-art performance in two human activity recognition tasks.
In Table~\ref{tab: performance_2}, we demonstrate the model efficiency of BioMamba against seven baselines across all proposed datasets. We found that Mamba-based models consistently show better learning efficiency and benefits from the selective state space mechanism with linear complexity. In the ADFTD~\cite{miltiadous2023dataset, miltiadous2023dice} and PTB-XL~\cite{wagner2020ptb} tasks, BioMamba incurs a high computational cost due to the dense setting of the frequency resolution.

\subsection{Visualization}
To visualize the effectiveness of BioMamba in further experiments, we depict the learned representation $\boldsymbol{Z^{''}}$ from the BioMamba Block, which setup on the UCI-HAR dataset~\cite{anguita2013public} as a case study. To visualize the representations more interpretably, we employ UMAP~\cite{mcinnes2018umap}, a dimensionality reduction technique with 25 neighbors and a minimum distance
of 0.5. To establish a reference standard, we utilize Mamba~\cite{gu2023mamba} and TCN~\cite{bai2018empirical}, since Mamba offers a novel architecture compared to attention-based and CNN-based methods, and TCN shows the best performance among other baselines.

\begin{figure*}[htpb]
    \centering
    \includegraphics[width=1\linewidth]{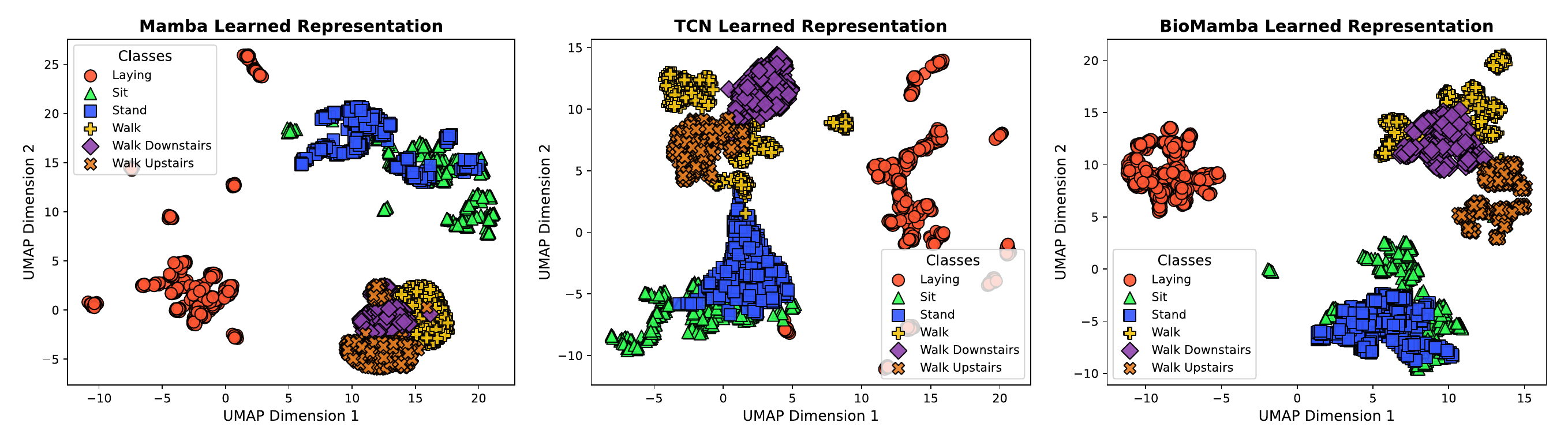} 
    \caption{Visualizing the learned representations from Mamba~\cite{gu2023mamba}, Medformer~\cite{wang2024medformer}, and our BioMamba. The visualized representations were trained from the encoder blocks on the UCI-HAR dataset~\cite{anguita2013public}. We present six different human activity representations, including Laying, Sit, Stand, Walk, Walk Upstairs, and Walk Downstairs. The results indicate that our approach more effectively segregates the six classes.}
    \label{fig: HAR_dataset}
\end{figure*}

\end{document}